\RequirePackage{fix-cm}
\documentclass[10pt,a4paper]{sensenova}

\usepackage{mtedit}

\title{MTEdit: Packed Trajectories Beat Rollouts\\
for Multi-Turn Image Editing}
\author{\parbox{\textwidth}{\centering\authorfont
  \href{https://github.com/houyuanchen111}{Houyuan Chen}\textsuperscript{\ensuremath{1,2,*}},
  \href{https://scholar.google.com/citations?user=dtXIEXoAAAAJ\&hl=zh-CN}{Fanzhou Wang}\textsuperscript{\ensuremath{1,*}},
  \href{https://x-niper.github.io/}{Xiangyu Fan}\textsuperscript{\ensuremath{1,\dagger}},
  \href{https://songlin1998.github.io/}{Songlin Yang}\textsuperscript{2},
  \href{https://scholar.google.com/citations?user=WMVmXgoAAAAJ}{Zhitao Yang}\textsuperscript{1}\\[2pt]
  \href{https://lllyasviel.github.io/lvmin_zhang/}{Lvmin Zhang}\textsuperscript{3},
  \href{https://scholar.google.com/citations?user=muMqKAcAAAAJ\&hl=zh-CN}{Pei Yang}\textsuperscript{\ensuremath{1,2}},
  \href{https://sites.google.com/view/fromandto}{Hao Zhao}\textsuperscript{\ensuremath{4,\ddagger}},
  \href{https://dahua.site/}{Dahua Lin}\textsuperscript{1},
  \href{https://yanglei.me/}{Lei Yang}\textsuperscript{\ensuremath{1,\ddagger}},
  \href{https://anyirao.com/}{Anyi Rao}\textsuperscript{\ensuremath{2,\ddagger}}
}}
\contribution[*]{Equal contribution}
\contribution[\dagger]{Project lead}
\contribution[\ddagger]{Co-corresponding authors}
\affiliation[1]{SenseTime Research}
\affiliation[2]{HKUST MMLab}
\affiliation[3]{Stanford University}
\affiliation[4]{AIR, Tsinghua University}
\hypersetup{pdfauthor={Houyuan Chen, Fanzhou Wang, Xiangyu Fan, Songlin Yang, Zhitao Yang, Lvmin Zhang, Pei Yang, Hao Zhao, Dahua Lin, Lei Yang, Anyi Rao}}
\abstract{
Multi-turn image editing aims to modify an image according to a sequence of instructions, making it better aligned with human-AI collaborative creation than single-turn editing.
However, constructing high-quality multi-turn training data remains challenging, as conventional sequential rollout accumulates errors and introduces visual-quality drift across editing turns.
To address this challenge, we introduce \textbf{MTEdit}, a multi-turn image editing dataset built through a pipeline centered on \textbf{packed trajectory canvas (PTC)} generation.
By jointly generating multiple editing states within a unified visual context, this pipeline reduces visual-quality drift across turns.
Covering both step-wise and context-aware editing scenarios, MTEdit comprises 500K multi-turn image editing trajectories, with up to 8 turns and an average of 5.38 turns per trajectory, making it one of the most extensive datasets for multi-turn image editing.
We further introduce \textbf{MTEdit-Bench}, a benchmark designed to evaluate both scenarios by assessing models' ability to follow editing instructions, maintain content consistency, resist visual-quality drift, and effectively use historical context across turns.
Using MTEdit, we fine-tune Qwen-Image-Edit-Causal to obtain \textbf{Qwen-Image-MTEdit}.
On MTEdit-Bench, we evaluate a broad range of closed-source and open-source image editing models, including Qwen-Image-MTEdit.
Experimental results show that Qwen-Image-MTEdit achieves the best performance among the evaluated open-source models, demonstrating the effectiveness of MTEdit in improving models' multi-turn image editing capabilities.
}

\begin{document}
\maketitle
\begin{figure}[H]
\centering
    
    \includegraphics[width=0.88\linewidth]{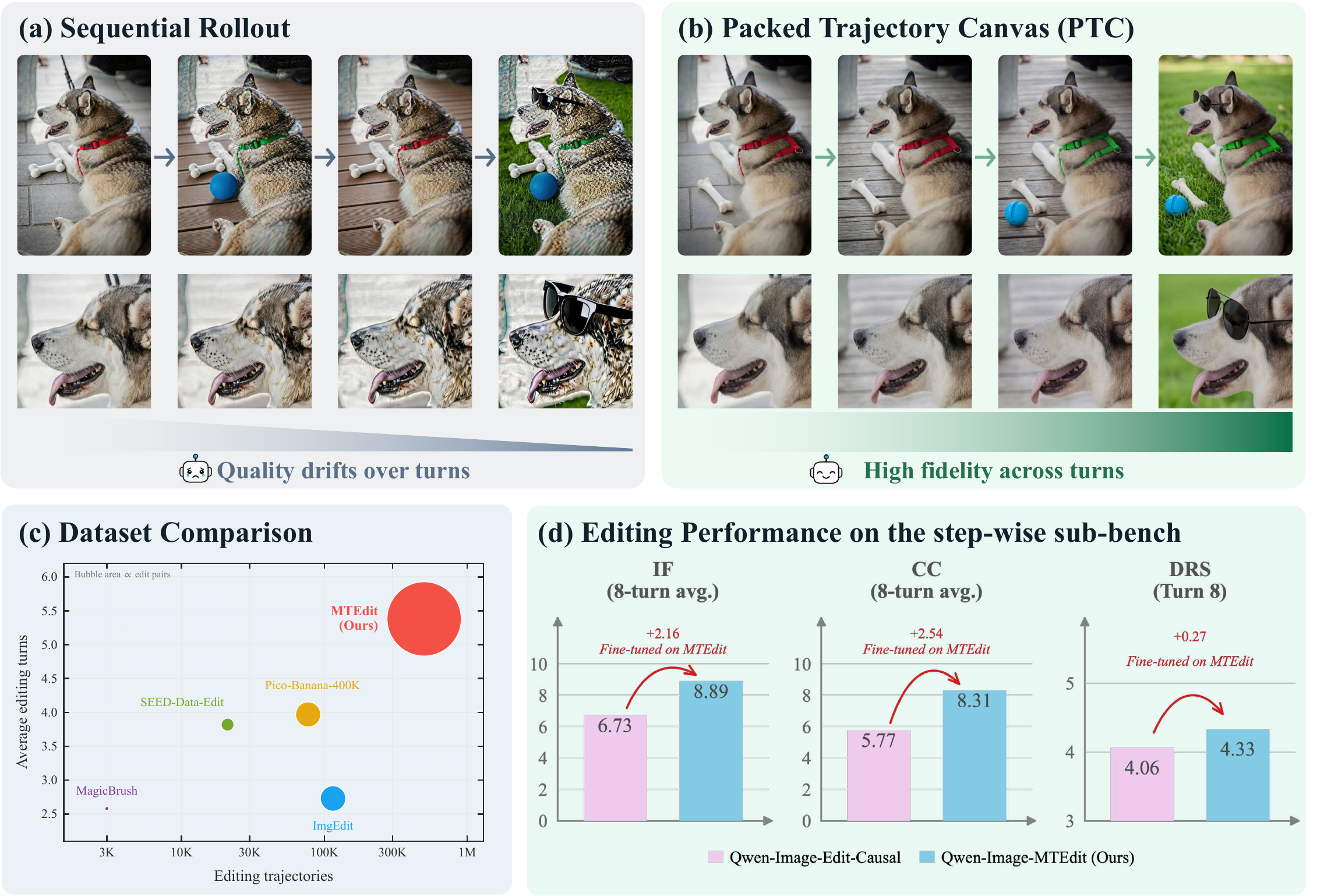}
    
    \caption{\textbf{Overview of our approach.} (a,b)~Packed trajectory canvas~(PTC) generation reduces visual-quality drift compared with sequential rollout. (c)~MTEdit offers more trajectories and turns than existing datasets. (d)~Fine-tuning on MTEdit improves multi-turn editing performance in instruction following~(IF), content consistency~(CC), and visual-quality stability~(DRS).}
    \label{fig:teaser}
\end{figure}

\clearpage

\section{Introduction}
\label{intro}

Recently, multi-turn image editing has attracted increasing attention from the research community.
This setting can be broadly categorized into two types: 1) \textit{step-wise} editing~\cite{zhou2025multi,liao2025freqedit,ye2026edit,ye2026agent}, where each editing step is performed independently based on the previous output image and the current instruction; and 2) \textit{context-aware} editing~\cite{ye2026imgedit,qu2026vincie,deng2025bagel,zhang2026non}, where the model generates the target image by jointly considering previous text instructions and historically generated images within a continuous editing context.
Compared with conventional single-turn editing~\cite{brooks2023instructpix2pix,sheynin2024emu,huang2024smartedit,li2025multiedit,mao2025visual,cui2025emu35nativemultimodalmodels,zhang2026enabling}, multi-turn image editing better aligns with the iterative workflow of real-world creative scenarios, suggesting substantial potential for practical applications.

Realizing this potential requires high-quality training data for both editing scenarios.
Sequential rollout with powerful closed-source image editing models is the most common approach to constructing such data, as it combines single-turn edits with a scalable generation process~\cite{qian2025pico}.
However, this approach reuses each generated image as the input to the next edit, allowing errors and artifacts to accumulate and causing visual-quality drift over longer trajectories.
Beyond visual quality, existing datasets are also limited in trajectory length and often cover only one of the two editing scenarios~\cite{zhang2023magicbrush,ye2026imgedit,qu2026vincie}.

To address these challenges, we introduce \textbf{MTEdit}, a multi-turn image editing dataset constructed through a pipeline centered on \textbf{packed trajectory canvas (PTC)} generation.
A PTC contains multiple panels, each depicting the outcome of one editing turn.
Jointly generating these panels within a unified visual context reduces visual-quality drift across turns, enabling the construction of high-quality multi-turn trajectories.
Specifically, given a curated source image, a vision-language model~(VLM) plans an editing sequence and produces a corresponding \textit{PTC prompt}.
The source image and this prompt are then provided to a closed-source image editing model~(e.g., GPT Image 2~\cite{openai2026gptimage2} or Nano Banana 2~\cite{google2026nanobanana2}) to generate the PTC.
Gemini 3.1 Pro~\cite{gemini31pro_modelcard} then filters out failed PTC generations before deriving turn-level editing instructions from each retained PTC and its source image.
The retained canvases are subsequently split into target images to form training pairs.
Finally, we apply SIFT-based~\cite{brown2007automatic} feature matching followed by image warping to improve pixel-level alignment within each editing pair.

MTEdit comprises a total of 500K multi-turn editing trajectories and 2.69M edit pairs, with 4 to 8 turns per trajectory and an average of 5.38 turns.
It covers both step-wise and context-aware editing scenarios and supports 23 common editing categories~(see Fig.~\ref{fig:dataset_statistics}), making it an extensive resource for training multi-turn image editing models.
As shown in Tab.~\ref{tab:comparison_of_multi_turn_image_edit_datasets} and Fig.~\ref{fig:teaser}(c), MTEdit offers more trajectories and longer editing sequences than the compared datasets.
Formal definitions of step-wise and context-aware editing are provided in Sec.~\ref{sec:method:task_definition}.

Beyond training data, reliable evaluation is essential for understanding model performance across successive editing turns.
Although existing benchmarks~\cite{qu2026vincie,chen2025edival,ye2026edit,xu2026anchoreditmaintainingtemporalconsistency} have begun to assess multi-turn image editing, many use sequences of only around three turns~\cite{ye2026imgedit,zhang2023magicbrush,ye2026edit,chen2025edival}, limiting their ability to reveal error accumulation over longer trajectories.
Their evaluation protocols also often overlook visual-quality drift across turns.
To address these limitations, we introduce \textbf{MTEdit-Bench} for evaluating both step-wise and context-aware editing.
It assesses instruction following, content consistency, resistance to visual-quality drift, and the ability to use historical instructions and image states throughout the editing process.

To demonstrate the training utility of MTEdit, we fine-tune Qwen-Image-Edit-Causal~\cite{modeltc2026qwenimageeditcausal} on the dataset to obtain \textbf{Qwen-Image-MTEdit}.
We evaluate a broad range of closed-source and open-source image editing models on MTEdit-Bench, including Qwen-Image-MTEdit.
Our model achieves the best performance among the evaluated open-source models, demonstrating the effectiveness of MTEdit in improving multi-turn image editing capabilities.

Our main contributions are summarized as follows: 1)~We introduce \textbf{MTEdit}, comprising 500K multi-turn editing trajectories constructed through a \textbf{PTC} generation pipeline that reduces visual-quality drift across turns; 2)~We introduce \textbf{MTEdit-Bench} to evaluate both step-wise and context-aware editing in terms of instruction following, content consistency, visual-quality stability, and historical context use; 3)~We develop \textbf{Qwen-Image-MTEdit} by fine-tuning on MTEdit, achieving the best performance among the evaluated open-source models on MTEdit-Bench.

\begin{table*}[t]
    \centering
    \caption{\textbf{Comparison of representative multi-turn image editing datasets.} \#Traj. and \#Edits denote the numbers of editing trajectories and edit pairs; Max T. and Avg T. denote the maximum and average numbers of turns; Step-wise and Context-aware indicate whether the dataset supports the two multi-turn editing settings; Construction describes how the dataset is built.}
    \label{tab:comparison_of_multi_turn_image_edit_datasets}
    \small
    \setlength{\tabcolsep}{4pt}
    \renewcommand{\arraystretch}{1.15}
    \resizebox{\textwidth}{!}{\begin{tabular}{lccccccl}
        \toprule
        Dataset & \#Traj. & \#Edits & Max T. & Avg T. & Step-wise & Context-aware & Construction \\
        \midrule
        MagicBrush~\cite{zhang2023magicbrush}   & 3K & 8K  & 3 & 2.58 & \checkmark & \ding{55}   & Human \\
        SEED-Data-Edit~\cite{ge2024seed}        & 21K  & 81K   & 6 & 3.82 & \checkmark & \ding{55}   & Human \\
        ImgEdit~\cite{ye2026imgedit}            & 115K & 314K  & 3 & 2.73 & \ding{55}  & \checkmark  & Inpainting \\
        Pico-Banana-400K~\cite{qian2025pico}    & 77K  & 306K  & 6 & 3.97 & \checkmark & \ding{55}   & Sequential rollout \\
        \midrule
        \textbf{MTEdit (Ours)} & \textbf{500K} & \textbf{2.69M} & \textbf{8} & \textbf{5.38} & \checkmark & \checkmark & \textbf{Packed trajectory canvas} \\
        \bottomrule
    \end{tabular}}
    
\end{table*}

\section{Related Work}
\label{related_work}

\paragraph{Instruction-based Image Editing:}
\label{related:iie}
Instruction-based image editing~(IIE) modifies an input image according to a natural language command while preserving instruction-irrelevant visual content.
InstructPix2Pix~\cite{brooks2023instructpix2pix} is the first work to formalize this paradigm by leveraging Prompt-to-Prompt~\cite{hertz2022prompt} to ensure spatial-structure consistency in training data.
Recent closed-source models such as the GPT-Image series~\cite{openai2025gptimage1,openai2025gptimage15,openai2026gptimage2} and Nano Banana series~\cite{google2025nanobanana,google2025nanobananapro,google2026nanobanana2}, together with open-source models such as Step1X-Edit~\cite{liu2025step1x}, FLUX~\cite{labs2025flux}, Qwen-Image-Edit~\cite{wu2025qwenimagetechnicalreport}, LongCat-Image~\cite{LongCat-Image}, FireRed-Image-Edit~\cite{firered2026rededit}, and JoyAI-Image~\cite{song2026awakingspatialintelligenceunified}, have continued to advance the field. The progress in this field largely depends on large-scale, high-quality training data~\cite{meng2021sdedit,sheynin2024emu,bai2024humanedit,fu2024guiding,zhang2024hive,chen2025opengpt,zhang2026weedit,chen2026scaleedit}.
However, most efforts still focus on single-turn editing, leaving the more interactive multi-turn setting comparatively underexplored.

\paragraph{Datasets for Single-turn Image Editing:}
\label{related:dataset_for_single_turn_image_edit}
Most existing image editing datasets are constructed under the single-turn editing setting.
Early datasets such as GIER~\cite{shi2021learning} provide human-annotated editing instructions with paired before-and-after images, establishing a standard supervision format for instruction-guided image editing.
Subsequent datasets scale this paradigm through synthetic or hybrid construction pipelines, including UltraEdit~\cite{zhao2024ultraedit}, OmniEdit~\cite{wei2025omniedit}, AnyEdit~\cite{yu2025anyedit}, and UniVG~\cite{fu2025univg}, while frontier-model-distilled datasets such as Echo-4o-Image~\cite{ye2025echo}, HQ-Edit~\cite{hui2024hq}, GPT-Image-Edit-1.5M~\cite{wang2025gpt} and ShareGPT-4o-Image~\cite{chen2025sharegpt4oimg} further expand the scale and complexity of image editing supervision.
Despite their effectiveness, these datasets mainly supervise one-step transformations, motivating dedicated datasets that capture coherent editing sequences across multiple successive user interaction turns rather than isolated before-and-after image pairs.

\paragraph{Datasets for Multi-turn Image Editing:}
MagicBrush~\cite{zhang2023magicbrush} provides early multi-turn samples for real images, but its trajectories are short, small-scale, and heavily human-annotated.
SEED-Data-Edit~\cite{ge2024seed} introduces Photoshop-expert annotations, yet its multi-turn subset remains limited and lacks systematic coverage of context-dependent editing.
ImgEdit~\cite{ye2026imgedit} explores context-aware abilities such as content understanding, content memory, and version backtracking, but it has short editing trajectories and overlooks common step-wise editing data.
Pico-Banana-400K~\cite{qian2025pico} constructs about 77K multi-turn sequences using Nano Banana, but its rollout pipeline can accumulate errors and introduce visual drift in longer trajectories.
VINCIE~\cite{qu2026vincie} extracts interleaved image-text sequences from videos for in-context editing, but natural video transitions are insufficient to cover diverse user editing intents.
Overall, high-quality, comprehensive, and large-scale datasets for multi-turn image editing remain lacking.

\begin{figure}[t]
    \centering
    \includegraphics[width=\linewidth]{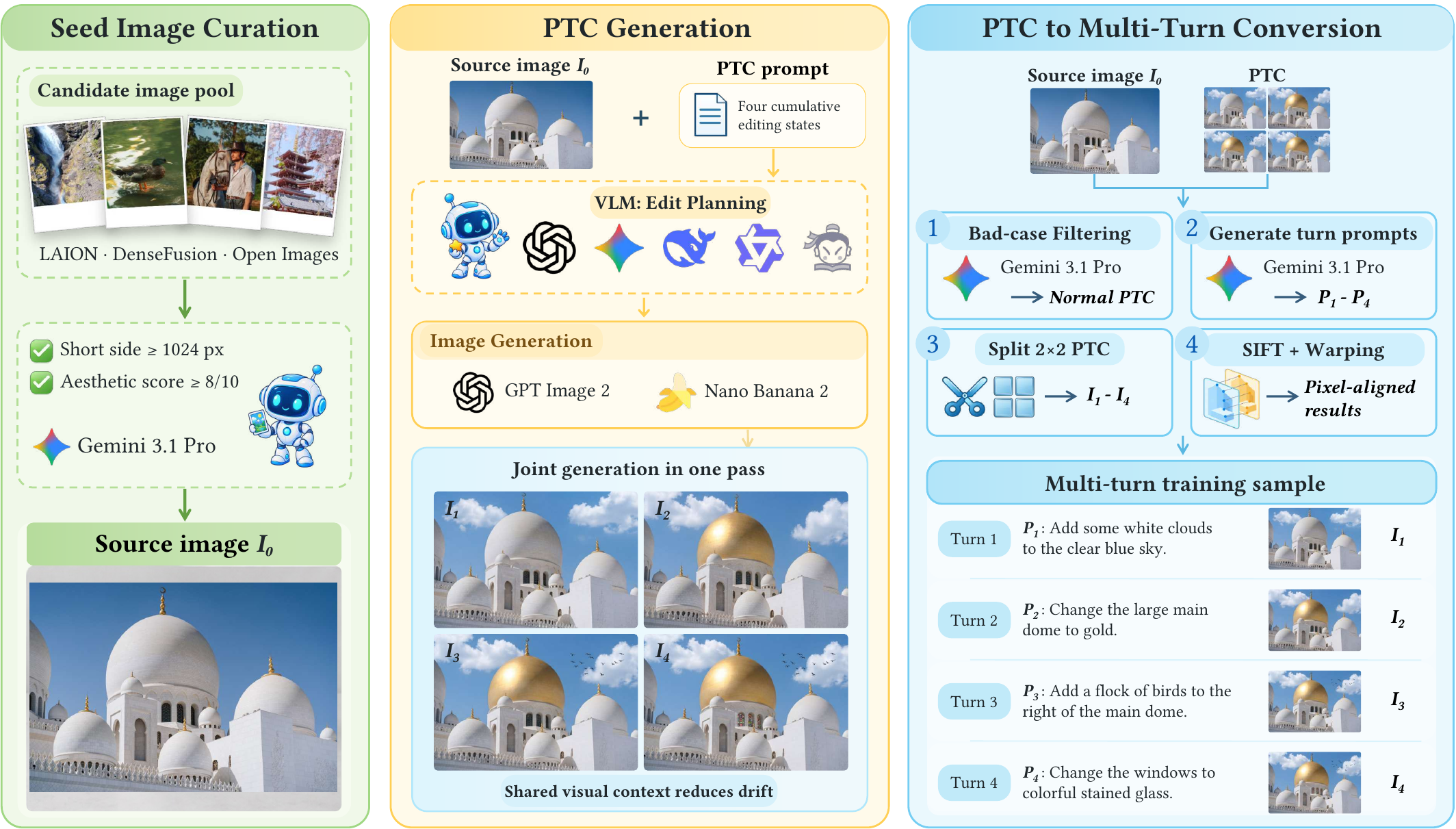}
    
    \caption{\textbf{Overview of our data construction pipeline.} (a)~Seed image curation, which selects source images with high resolution and aesthetic quality; (b)~PTC generation, in which VLMs derive a PTC prompt from each source image to guide closed-source image models in generating the corresponding PTC; and (c)~PTC to multi-turn conversion, which processes the generated canvases into multi-turn editing trajectories with turn-level instructions and target images.}
    \label{fig:data_construction_pipeline}
    
\end{figure}

\section{Technical Details}
\label{sec:method}
This section presents the technical details of MTEdit, MTEdit-Bench, and Qwen-Image-MTEdit.
Sec.~\ref{sec:method:task_definition} first defines the tasks covered by our dataset: \textit{step-wise} editing and three types of \textit{context-aware} editing.
Sec.~\ref{sec:method:data_construction_pipeline} details the data construction pipeline of MTEdit.
Sec.~\ref{sec:method:dataset_statistics} analyzes the scale of MTEdit, the distribution of editing types, and its coverage of multi-turn scenarios.
Sec.~\ref{sec:method:mt_bench} introduces the construction of MTEdit-Bench and its automated evaluation protocol.
Sec.~\ref{sec:method:qwen_image_edit_mt} further describes Qwen-Image-MTEdit, which is fine-tuned on MTEdit.

\subsection{Task Definition}
\label{sec:method:task_definition}

\paragraph{Step-wise Editing} is the most common setting in multi-turn image editing and is also the primary form studied in most prior work~\cite{zhou2025multi,liao2025freqedit,ye2026edit}. In this setting, each editing turn does not explicitly rely on the historical context composed of all previous images and text instructions. Instead, it is mainly conditioned on the image generated in the previous turn and the current user edit prompt.
Therefore, models designed for single-turn editing can also be extended to step-wise editing through rollout-style execution. However, for single-turn edit models, this strategy causes errors to accumulate across editing turns. As the number of turns increases, image content, subject consistency, and local details often exhibit noticeable drift~\cite{tang2026banana100}.

\paragraph{Context-aware Editing} requires the model to leverage historical information, rather than relying solely on the latest image and current instruction. We consider three context-aware editing types: (1) \textit{persistent-constraint} editing, where constraints specified in earlier prompts must remain valid in subsequent turns; (2) \textit{textual-reference} editing, where later instructions refer to previously introduced or modified objects through pronouns or descriptive phrases; and (3) \textit{short-instruction} editing, where an initial explicit instruction establishes the edit target and later-turn instructions are brief or underspecified, often requesting adjustments to the degree or intensity of an attribute.

\subsection{Data Construction Pipeline}
\label{sec:method:data_construction_pipeline}
The data construction pipeline of MTEdit consists of three steps: (a) seed image curation, (b) PTC generation, and (c) PTC to multi-turn conversion, as illustrated in Fig.~\ref{fig:data_construction_pipeline}.

\paragraph{Seed Image Curation:}
To construct multi-turn editing trajectories, we first curate seed images that serve as the source images of the trajectories. We collect candidate images from large-scale, openly available image sources~\cite{chen2025janus}, including LAION~\cite{schuhmann2022laion}, Open Images~\cite{kuznetsova2020open} and DenseFusion~\cite{li2024densefusion}. Each retained seed image is used as the initial state of one multi-turn editing trajectory.

To ensure that the source images have high visual quality, we apply two filtering criteria: (i) high resolution, requiring the shorter side of the image to be at least 1024 pixels; and (ii) high aesthetic quality, requiring an aesthetic score of at least $8$ on a $0$--$10$ scale, as assessed by Gemini 3.1 Pro~\cite{gemini31pro_modelcard}. The system prompt is in Appendix~\ref{app:sys_prompt_of_seed_image_curation}.

\paragraph{PTC Generation:}
A common strategy~\cite{qian2025pico} for constructing multi-turn image editing trajectories is to leverage the capabilities of closed-source image editing models and generate data in a sequential rollout manner, where the output of each editing turn is used as the input to the next turn. However, the major limitation of this strategy is that visual quality drift becomes increasingly severe as the number of editing turns grows (e.g., beyond four turns).

To address this issue, we use closed-source models to generate multiple editing states within a single PTC in one pass. By jointly generating targets for multiple editing turns in a shared context, this approach reduces visual quality drift and enables longer editing trajectories with better visual quality. Qualitative and quantitative comparisons with rollout-style pipelines are presented in Fig.~\ref{fig:drift_comparison} and Tab.~\ref{tab:drift_comparison} in the appendix, respectively. PTC generation proceeds in two steps. \textbf{First}, we use VLMs to synthesize an instruction for closed-source models from the source image, named a \textit{PTC prompt}. \textbf{Second}, we feed the source image and \textit{PTC prompt} to closed-source models to obtain the PTC. The generated PTCs are 4K square canvases with four panels in a $2{\times}2$ layout, ensuring that each panel retains a resolution of approximately 2K, which is sufficient for producing high-quality training data.

\paragraph{PTC to Multi-Turn Conversion:}
After obtaining the PTC, we convert it into multi-turn image editing trajectories. This conversion consists of four steps.
\textbf{First}, we use Gemini 3.1 Pro to filter out failed PTC generations, such as those shown in Fig.~\ref{fig:grid_image_failure_case}.
\textbf{Second}, we feed the source image with the retained PTC into Gemini 3.1 Pro to generate user prompts for each editing turn.
\textbf{Third}, we split the PTC into panels according to the $2{\times}2$ layout, and treat each panel as the target image after the corresponding editing turn.
\textbf{Fourth}, we apply SIFT-based~\cite{brown2007automatic} feature matching followed by image warping to improve pixel-level alignment within each editing pair.
As a result, each sample consists of the source image, turn-level user prompts, and turn-level target images.

Moreover, under the default $2{\times}2$ PTC layout, each run of our data construction pipeline produces a single four-turn editing sample. To construct longer editing sequences~(such as eight-turn samples), we only need to take the generated image from the last turn as the new source image for the next round and rerun the pipeline.

\begin{figure*}[t]
    \centering
    \includegraphics[width=\textwidth]{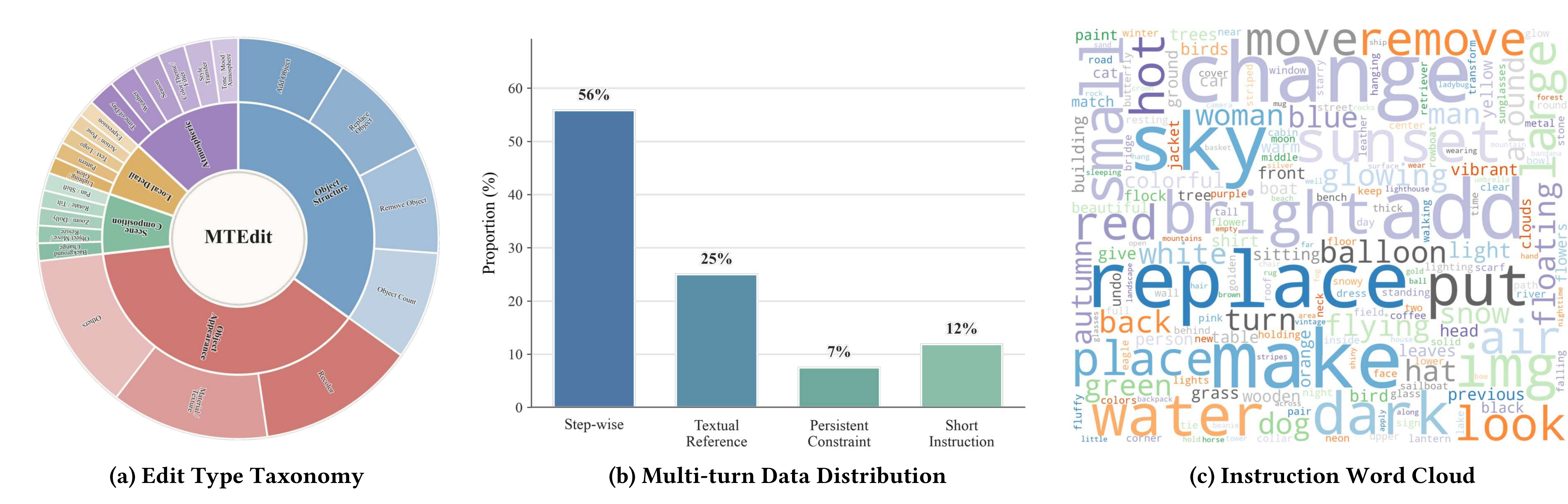}
    
    \caption{\textbf{Statistics of MTEdit.} (a)~Edit-type taxonomy, (b)~multi-turn data distribution, and (c)~instruction word cloud.}
    \label{fig:dataset_statistics}
    
\end{figure*}

\subsection{Dataset Statistics}
\label{sec:method:dataset_statistics}
MTEdit contains 500K editing trajectories and 2.69M edit pairs, with up to 8 turns and an average of 5.38 turns per trajectory, and simultaneously covers both step-wise and context-aware multi-turn editing. In terms of the number of turns, the total number of edit pairs, and scenario completeness, our dataset provides broader coverage than the compared multi-turn editing datasets.

Fig.~\ref{fig:dataset_statistics} further characterizes the distribution of MTEdit from three perspectives. The left subfigure summarizes the edit-type taxonomy used in MTEdit. We organize all edits into a two-level hierarchy, with five coarse categories: object structure, object appearance, scene composition, local detail, and atmospheric edits. These coarse categories are further divided into 23 leaf categories.
The middle subfigure reports the proportion of each multi-turn data type: step-wise data accounts for 56\%, while the remaining 44\% are context-aware data, further divided into textual-reference~(25\%), persistent-constraint~(7\%) and short-instruction~(12\%). The right subfigure presents a word cloud of all user instructions in MTEdit, illustrating its lexical diversity.

\begin{figure}[t]
    \centering
    \includegraphics[width=\linewidth]{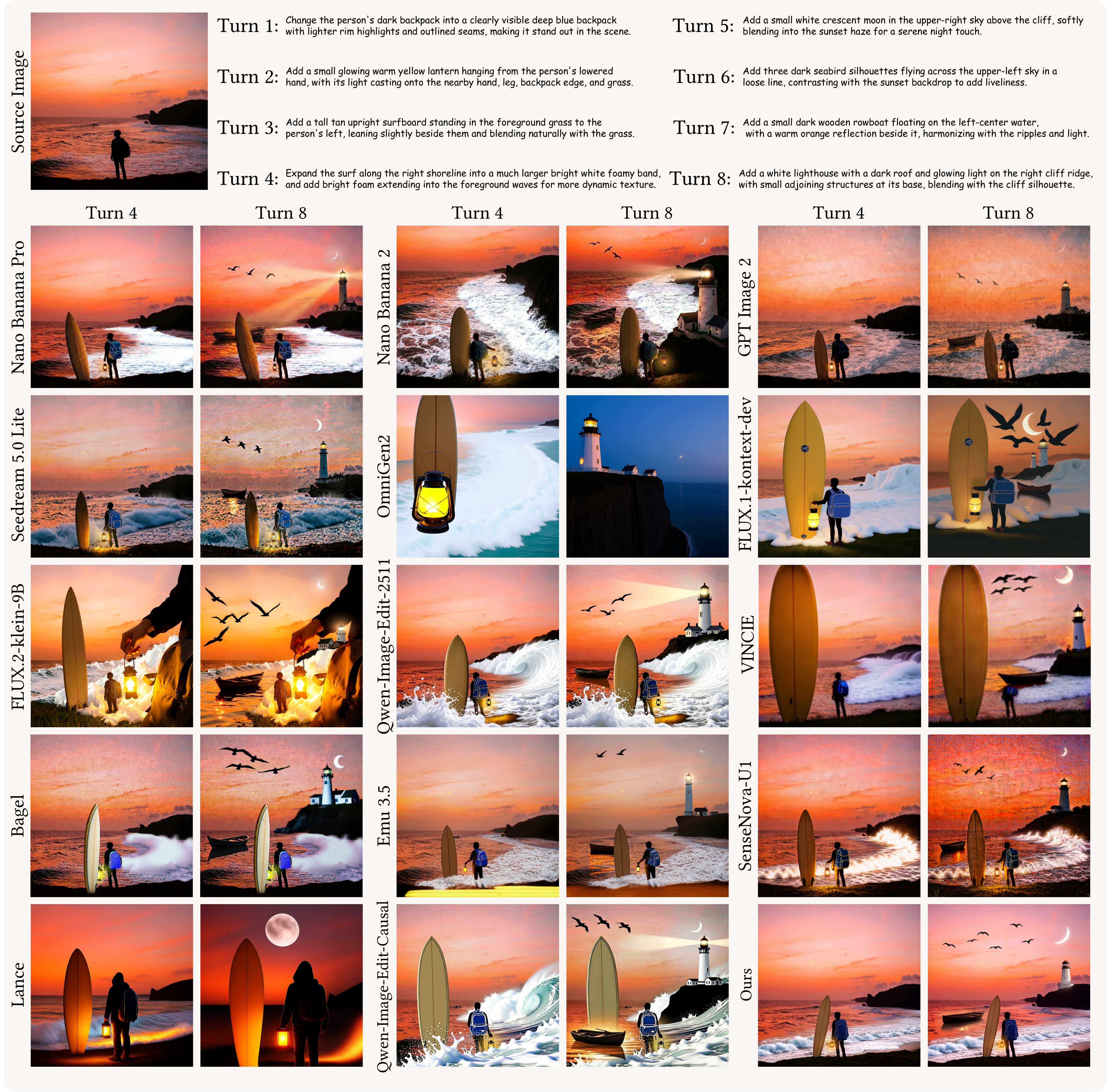}

\caption{\textbf{Qualitative results on the step-wise sub-bench.} Our method achieves competitive editing performance and maintains visual quality without noticeable degradation across editing turns.}
    \label{fig:step_wise_results}

\end{figure}

\subsection{MTEdit-Bench}

\label{sec:method:mt_bench}

\paragraph{Benchmark Construction and Task Coverage:} MTEdit-Bench comprises source images paired with manually crafted prompts, covering four multi-turn editing tasks: step-wise, persistent-constraint, textual-reference, and short-instruction. The step-wise sub-bench consists of 100 eight-turn samples and mainly evaluates editing performance and visual drift over long editing chains. Each of the three context-aware sub-benches consists of 50 four-turn samples and assesses models’ use of prior textual prompts and visual states across turns.

\paragraph{Evaluation Metrics:}
MTEdit-Bench uses task-specific metrics to evaluate capabilities required across multi-turn editing scenarios.
For the step-wise sub-bench, we report three turn-level metrics: \(\mathrm{IF}\) (Instruction Following), \(\mathrm{CC}\) (Content Consistency), and \(\mathrm{DRS}\) (Drift Robustness Score).
\(\mathrm{IF}\) and \(\mathrm{CC}\) follow the VLM-judge-based evaluation paradigm used in VIEScore~\cite{ku2024viescore}, GEdit-Bench~\cite{liu2025step1x}, Edit-R2~\cite{ye2026edit}, and EdiVal-Agent~\cite{chen2025edival}, measuring instruction following and content consistency. To compute \(\mathrm{IF}\) at each turn, we provide the VLM with the pre-edit image, the model's generated image after the edit, and the current-turn instruction. To compute \(\mathrm{CC}\), we provide the initial image of the editing sequence, the model's generated image at the current turn, and all instructions up to that turn. The VLM assigns each metric a score from 0 to 10~(see Appendix~\ref{sec:sys_prompt} for the system prompts). If either \(\mathrm{IF}\) or \(\mathrm{CC}\) falls below 6 at any turn, we consider the editing trajectory to have failed and set both scores to 0 for all subsequent turns. This reflects the causal nature of multi-turn editing: subsequent edits build on an incorrect state after a failure and are therefore no longer credited as successful continuation, consistent with the failure-propagation principle used in VINCIE~\cite{qu2026vincie}. To balance performance and reproducibility, we use Qwen 3.5 Plus~\cite{qwen35blog} as the VLM evaluator.

\(\mathrm{DRS}\) measures robustness against visual drift and is computed from two drift-sensitive metrics, VisualQuality-R1~\cite{wu2026visualquality} and RALI~\cite{zhao2025reasoning}. \(\mathrm{DRS}\) is calculated as the average of their image quality scores, \((\mathrm{VisualQuality\text{-}R1}+\mathrm{RALI})/2\), and ranges from 0 to 5. This choice is motivated by Banana100~\cite{tang2026banana100}, which finds that VisualQuality-R1 and RALI reliably detect visual quality degradation during iterative image replication, assigning lower scores to degraded images. We use \(\mathrm{DRS}\) to evaluate all types of multi-turn editing data.

For the remaining three types of context-aware editing, it is necessary to consider whether models can correctly use the historical context. We therefore provide the VLM judge with the full multi-turn editing trajectory, including the instructions and corresponding generated images, allowing it to assess each case within the shared interaction context. We denote the resulting instruction-following and content-consistency scores as \(\mathrm{IF}^{\ast}\) and \(\mathrm{CC}^{\ast}\), respectively. The same failure-propagation rule applies to these two scores; \(\mathrm{DRS}\) is never zeroed after a failure.

\subsection{Qwen-Image-MTEdit}
\label{sec:method:qwen_image_edit_mt}

To assess the training utility of MTEdit, we fine-tune Qwen-Image-Edit-Causal~\cite{modeltc2026qwenimageeditcausal} on this data and obtain \textbf{Qwen-Image-MTEdit}.
Building on the block causal attention, we adopt a teacher-forcing training strategy~\cite{chen2024diffusion}, which allows multiple editing turns within the same trajectory to be trained jointly rather than through sequential rollout.
In addition, the model uses 3D interleaved RoPE to encode the temporal order of interleaved image-text units, enabling it to distinguish different editing turns and align each instruction with its corresponding image state.
Further analysis and detailed training settings are provided in Appendix~\ref{app:qwen_image_edit_mt_details}.

\begin{table}[t]
    \centering
    \definecolor{closedsourcebg}{RGB}{242,243,245}
    \definecolor{opensourcebg}{RGB}{241,237,249}
    \definecolor{oursbg}{RGB}{226,241,241}
    \definecolor{bestgreen}{RGB}{200,225,223}
    \definecolor{secondgreen}{RGB}{232,243,241}
    \newcommand{\stepwiselegendcell}[2]{{\setlength{\fboxsep}{1.5pt}\colorbox{#1}{\strut #2}}}
\caption{\textbf{Quantitative results on the step-wise sub-bench.} Closed-source scores are shown in \textcolor{black!55}{gray}; among open-source methods and our model, different cell colors refer to \stepwiselegendcell{bestgreen}{best} and \stepwiselegendcell{secondgreen}{2nd-best} results. For all metrics, higher values are better.}
    \label{tab:step_wise_results}
    \footnotesize
    \setlength{\tabcolsep}{3pt}
    \renewcommand{\arraystretch}{1.15}
    \newcommand{\bestcell}[1]{\cellcolor{bestgreen}#1}
    \newcommand{\secondcell}[1]{\cellcolor{secondgreen}#1}
    \newcommand{\closedval}[1]{\textcolor{black!55}{#1}}
    \begin{tabular*}{\linewidth}{@{\extracolsep{\fill}}l*{12}{c}@{}}
        \toprule
        \multicolumn{1}{l}{}
        & \multicolumn{3}{c}{\textbf{Turn 1}}
        & \multicolumn{3}{c}{\textbf{Turn 2}}
        & \multicolumn{3}{c}{\textbf{Turn 3}}
        & \multicolumn{3}{c}{\textbf{Turn 4}} \\
        \cmidrule(lr){2-4}
        \cmidrule(lr){5-7}
        \cmidrule(lr){8-10}
        \cmidrule(lr){11-13}
        \textbf{Method} & \(\mathrm{IF}\) & \(\mathrm{CC}\) & \(\mathrm{DRS}\)
        & \(\mathrm{IF}\) & \(\mathrm{CC}\) & \(\mathrm{DRS}\)
        & \(\mathrm{IF}\) & \(\mathrm{CC}\) & \(\mathrm{DRS}\)
        & \(\mathrm{IF}\) & \(\mathrm{CC}\) & \(\mathrm{DRS}\) \\
        \midrule
        \cellcolor{closedsourcebg}Nano Banana Pro~\cite{google2025nanobananapro}
        & \closedval{9.75} & \closedval{9.79} & \closedval{4.17}
        & \closedval{9.73} & \closedval{9.53} & \closedval{4.22}
        & \closedval{9.63} & \closedval{9.35} & \closedval{4.22}
        & \closedval{9.53} & \closedval{9.06} & \closedval{4.19} \\
        \cellcolor{closedsourcebg}Nano Banana 2~\cite{google2026nanobanana2}
        & \closedval{9.65} & \closedval{9.58} & \closedval{4.19}
        & \closedval{9.45} & \closedval{9.21} & \closedval{4.21}
        & \closedval{9.28} & \closedval{8.85} & \closedval{4.23}
        & \closedval{9.31} & \closedval{8.53} & \closedval{4.20} \\
        \cellcolor{closedsourcebg}GPT Image 2~\cite{openai2026gptimage2}
        & \closedval{9.02} & \closedval{8.72} & \closedval{4.22}
        & \closedval{8.66} & \closedval{8.62} & \closedval{4.23}
        & \closedval{8.65} & \closedval{8.07} & \closedval{4.22}
        & \closedval{8.62} & \closedval{7.82} & \closedval{4.20} \\
        \cellcolor{closedsourcebg}Seedream 5.0 Lite~\cite{bytedance2026seedream5lite}
        & \closedval{9.73} & \closedval{9.60} & \closedval{4.27}
        & \closedval{9.81} & \closedval{9.08} & \closedval{4.29}
        & \closedval{9.66} & \closedval{8.57} & \closedval{4.26}
        & \closedval{9.34} & \closedval{7.69} & \closedval{4.19} \\
        \midrule
        \cellcolor{opensourcebg}VINCIE~\cite{qu2026vincie}
        & 7.24 & 7.05 & 3.28
        & 5.97 & 4.76 & 3.01
        & 4.19 & 3.05 & 2.86
        & 2.30 & 1.66 & 2.71 \\
        \cellcolor{opensourcebg}OmniGen2~\cite{wu2025omnigen2}
        & 6.75 & 7.46 & 4.00
        & 4.44 & 4.95 & 3.77
        & 3.02 & 2.75 & 3.43
        & 1.52 & 1.24 & 3.05 \\
        \cellcolor{opensourcebg}FLUX.1-kontext-dev~\cite{labs2025flux}
        & 8.29 & 8.77 & 4.12
        & 7.41 & 7.61 & 4.12
        & 6.40 & 6.42 & 4.12
        & 5.04 & 5.08 & 4.09 \\
        \cellcolor{opensourcebg}FLUX.2-klein-9B~\cite{flux-2-2025}
        & 9.09 & 9.19 & 4.15
        & 8.63 & 8.83 & 4.17
        & 8.23 & 8.03 & 4.21
        & 8.06 & \secondcell{7.53} & 4.20 \\
        \cellcolor{opensourcebg}Qwen-Image-Edit-2511~\cite{wu2025qwenimagetechnicalreport}
        & \secondcell{9.30} & \secondcell{9.43} & 4.12
        & \secondcell{9.07} & \secondcell{8.87} & 4.15
        & \secondcell{8.67} & \secondcell{8.11} & 4.16
        & \secondcell{8.31} & 7.51 & 4.16 \\
        \cellcolor{opensourcebg}BAGEL~\cite{deng2025bagel}
        & 8.69 & 9.14 & 3.99
        & 7.56 & 8.18 & 3.92
        & 6.48 & 6.95 & 3.88
        & 4.57 & 5.55 & 3.71 \\
        \cellcolor{opensourcebg}Emu3.5~\cite{cui2025emu35nativemultimodalmodels}
        & 7.40 & 7.90 & 4.03
        & 7.09 & 6.83 & 4.01
        & 6.17 & 5.98 & 4.16
        & 5.69 & 5.30 & 4.13 \\
        \cellcolor{opensourcebg}SenseNova-U1~\cite{sensenova2026sensenovau1}
        & 9.29 & 9.06 & 4.07
        & 8.95 & 8.51 & 4.14
        & 8.53 & 7.61 & 4.12
        & 7.37 & 5.95 & 3.90 \\
        \cellcolor{opensourcebg}Lance~\cite{fu2026lanceunifiedmultimodalmodeling}
        & 8.20 & 8.39 & \secondcell{4.18}
        & 6.78 & 6.25 & \bestcell{4.29}
        & 4.44 & 3.77 & \bestcell{4.32}
        & 2.59 & 1.78 & \secondcell{4.30} \\
        \midrule
        \cellcolor{oursbg}\shortstack[l]{Qwen-Image-Edit-Causal~\cite{modeltc2026qwenimageeditcausal}}
        & 8.89 & 9.21 & \secondcell{4.18}
        & 8.76 & 8.21 & 4.22
        & 8.54 & 7.26 & 4.20
        & 7.75 & 6.38 & 4.17 \\
        \cellcolor{oursbg}\textbf{Ours}
        & \bestcell{9.74} & \bestcell{9.54} & \bestcell{4.23}
        & \bestcell{9.59} & \bestcell{9.28} & \secondcell{4.26}
        & \bestcell{9.32} & \bestcell{9.19} & \secondcell{4.30}
        & \bestcell{9.35} & \bestcell{8.82} & \bestcell{4.33} \\
        \midrule[\heavyrulewidth]
        \addlinespace[0.8ex]
        \multicolumn{1}{l}{}
        & \multicolumn{3}{c}{\textbf{Turn 5}}
        & \multicolumn{3}{c}{\textbf{Turn 6}}
        & \multicolumn{3}{c}{\textbf{Turn 7}}
        & \multicolumn{3}{c}{\textbf{Turn 8}} \\
        \cmidrule(lr){2-4}
        \cmidrule(lr){5-7}
        \cmidrule(lr){8-10}
        \cmidrule(lr){11-13}
        \textbf{Method} & \(\mathrm{IF}\) & \(\mathrm{CC}\) & \(\mathrm{DRS}\)
        & \(\mathrm{IF}\) & \(\mathrm{CC}\) & \(\mathrm{DRS}\)
        & \(\mathrm{IF}\) & \(\mathrm{CC}\) & \(\mathrm{DRS}\)
        & \(\mathrm{IF}\) & \(\mathrm{CC}\) & \(\mathrm{DRS}\) \\
        \midrule
        \cellcolor{closedsourcebg}Nano Banana Pro~\cite{google2025nanobananapro}
        & \closedval{9.54} & \closedval{8.73} & \closedval{4.15}
        & \closedval{9.45} & \closedval{8.49} & \closedval{4.12}
        & \closedval{8.94} & \closedval{8.31} & \closedval{4.08}
        & \closedval{8.83} & \closedval{7.51} & \closedval{4.05} \\
        \cellcolor{closedsourcebg}Nano Banana 2~\cite{google2026nanobanana2}
        & \closedval{9.24} & \closedval{8.38} & \closedval{4.19}
        & \closedval{8.98} & \closedval{7.86} & \closedval{4.19}
        & \closedval{8.34} & \closedval{7.44} & \closedval{4.18}
        & \closedval{7.99} & \closedval{6.80} & \closedval{4.20} \\
        \cellcolor{closedsourcebg}GPT Image 2~\cite{openai2026gptimage2}
        & \closedval{8.58} & \closedval{7.32} & \closedval{4.18}
        & \closedval{8.23} & \closedval{7.00} & \closedval{4.16}
        & \closedval{7.41} & \closedval{6.26} & \closedval{4.17}
        & \closedval{7.11} & \closedval{5.81} & \closedval{4.17} \\
        \cellcolor{closedsourcebg}Seedream 5.0 Lite~\cite{bytedance2026seedream5lite}
        & \closedval{9.07} & \closedval{7.11} & \closedval{4.10}
        & \closedval{8.89} & \closedval{6.33} & \closedval{4.00}
        & \closedval{7.31} & \closedval{5.33} & \closedval{3.88}
        & \closedval{6.38} & \closedval{4.52} & \closedval{3.75} \\
        \midrule
        \cellcolor{opensourcebg}VINCIE~\cite{qu2026vincie}
        & 1.27 & 0.89 & 2.68
        & 0.59 & 0.42 & 2.57
        & 0.14 & 0.20 & 2.61
        & 0.05 & 0.06 & 2.59 \\
        \cellcolor{opensourcebg}OmniGen2~\cite{wu2025omnigen2}
        & 0.58 & 0.48 & 2.98
        & 0.31 & 0.21 & 2.95
        & 0.20 & 0.07 & 2.98
        & 0.00 & 0.00 & 3.10 \\
        \cellcolor{opensourcebg}FLUX.1-kontext-dev~\cite{labs2025flux}
        & 4.38 & 4.04 & 4.00
        & 3.62 & 3.20 & 3.95
        & 2.47 & 2.50 & 3.91
        & 2.13 & 1.76 & 3.82 \\
        \cellcolor{opensourcebg}FLUX.2-klein-9B~\cite{flux-2-2025}
        & 7.48 & \secondcell{7.12} & 4.20
        & 7.40 & \secondcell{6.58} & 4.17
        & 6.39 & \secondcell{6.23} & 4.18
        & 6.07 & 5.43 & \secondcell{4.12} \\
        \cellcolor{opensourcebg}Qwen-Image-Edit-2511~\cite{wu2025qwenimagetechnicalreport}
        & \secondcell{8.15} & 7.06 & 4.13
        & \secondcell{7.56} & 6.52 & 4.11
        & \secondcell{7.10} & 6.15 & 4.12
        & \secondcell{6.87} & \secondcell{5.56} & 4.05 \\
        \cellcolor{opensourcebg}BAGEL~\cite{deng2025bagel}
        & 3.99 & 4.14 & 3.62
        & 2.79 & 3.19 & 3.37
        & 2.15 & 2.17 & 3.24
        & 1.60 & 1.68 & 3.06 \\
        \cellcolor{opensourcebg}Emu3.5~\cite{cui2025emu35nativemultimodalmodels}
        & 5.15 & 4.74 & 4.13
        & 4.40 & 4.21 & 4.17
        & 3.48 & 3.17 & 4.06
        & 3.01 & 2.72 & \secondcell{4.12} \\
        \cellcolor{opensourcebg}SenseNova-U1~\cite{sensenova2026sensenovau1}
        & 6.21 & 4.41 & 3.55
        & 4.50 & 3.06 & 3.14
        & 2.96 & 1.96 & 2.84
        & 1.67 & 1.12 & 2.58 \\
        \cellcolor{opensourcebg}Lance~\cite{fu2026lanceunifiedmultimodalmodeling}
        & 0.66 & 0.58 & \secondcell{4.28}
        & 0.18 & 0.13 & \secondcell{4.24}
        & 0.00 & 0.03 & \secondcell{4.20}
        & 0.00 & 0.00 & 4.11 \\
        \midrule
        \cellcolor{oursbg}\shortstack[l]{Qwen-Image-Edit-Causal~\cite{modeltc2026qwenimageeditcausal}}
        & 6.87 & 5.20 & 4.15
        & 5.74 & 4.29 & 4.13
        & 4.11 & 3.21 & 4.12
        & 3.17 & 2.41 & 4.06 \\
        \cellcolor{oursbg}\textbf{Ours}
        & \bestcell{8.99} & \bestcell{8.12} & \bestcell{4.32}
        & \bestcell{8.70} & \bestcell{7.60} & \bestcell{4.33}
        & \bestcell{8.00} & \bestcell{7.51} & \bestcell{4.35}
        & \bestcell{7.45} & \bestcell{6.42} & \bestcell{4.33} \\
        \bottomrule
    \end{tabular*}
    
\end{table}

\section{Experiments}
\label{sec:experiments}

\subsection{Experimental Setup}
\label{sec:experiments:experimental_setup}

\paragraph{Benchmark and Metrics:}
The construction details and evaluation protocol of MTEdit-Bench can be found in Sec.~\ref{sec:method:mt_bench}. In brief, the step-wise sub-bench reports \(\mathrm{IF}\), \(\mathrm{CC}\), and \(\mathrm{DRS}\); the remaining three context-aware sub-benches report \(\mathrm{IF}^{\ast}\), \(\mathrm{CC}^{\ast}\), and \(\mathrm{DRS}\). All tables present turn-wise results, where each score at a turn is averaged over all test samples evaluated at that turn.

\paragraph{Compared Methods:}
We group the compared methods into two categories based on their ability to use historical image-text context.
The first category includes Nano Banana Pro~\cite{google2025nanobananapro}, Nano Banana 2~\cite{google2026nanobanana2}, GPT Image 2~\cite{openai2026gptimage2}, Seedream 5.0 Lite~\cite{bytedance2026seedream5lite}, BAGEL~\cite{deng2025bagel}, Emu3.5~\cite{cui2025emu35nativemultimodalmodels}, SenseNova-U1~\cite{sensenova2026sensenovau1}, Lance~\cite{fu2026lanceunifiedmultimodalmodeling}, VINCIE~\cite{qu2026vincie}, and our Qwen-Image-MTEdit. These methods can leverage historical images and instructions, supporting both step-wise and context-aware multi-turn editing.
The second category includes OmniGen2~\cite{wu2025omnigen2}, FLUX.1-kontext-dev~\cite{labs2025flux1kontextflowmatching}, FLUX.2-klein-9B~\cite{flux-2-2025}, Qwen-Image-Edit-2511~\cite{wu2025qwenimagetechnicalreport}, and Qwen-Image-Edit-Causal~\cite{modeltc2026qwenimageeditcausal}. These methods perform multi-turn editing by conditioning each turn on the previous output image and the current instruction. Since they do not explicitly model historical context, we evaluate them only in the step-wise setting.

\begin{figure}[t]
    \centering
    \includegraphics[width=\linewidth]{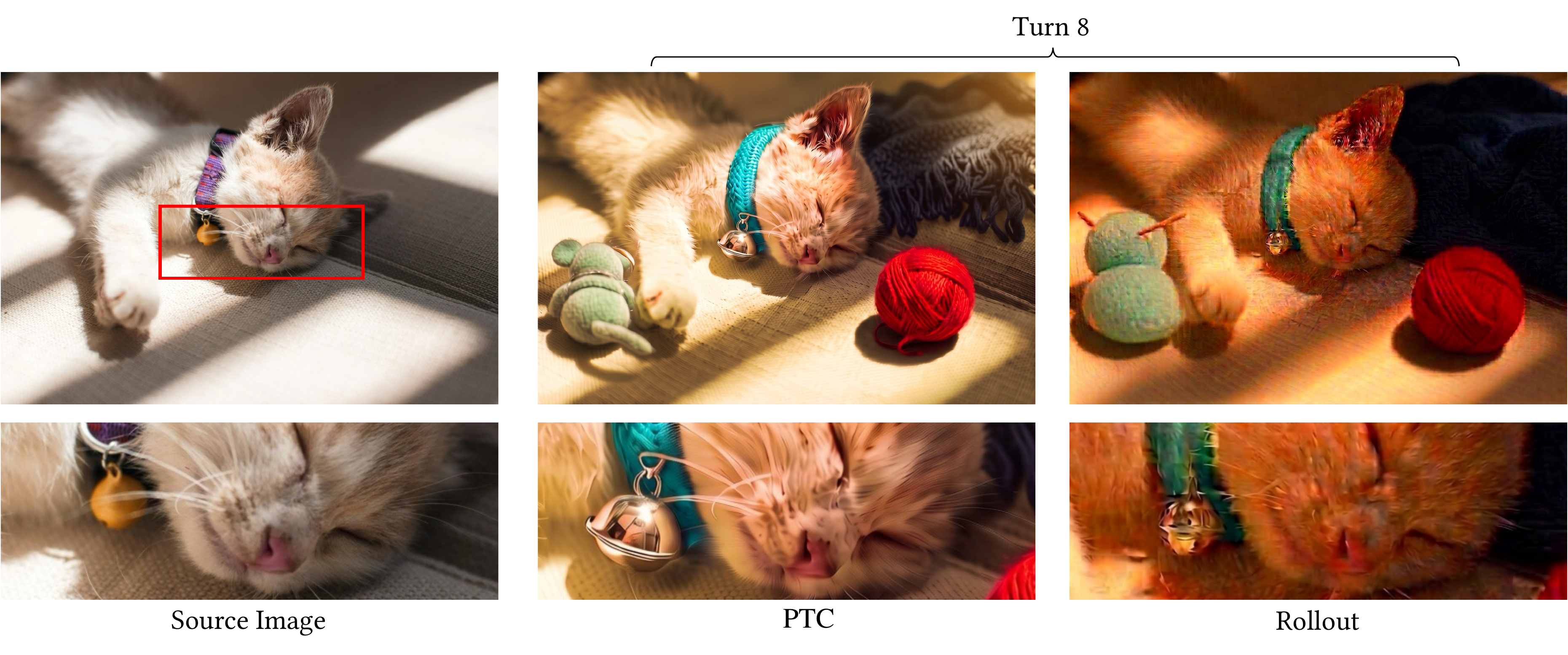}

\caption{\textbf{Visual quality comparison at Turn~8.} From left to right: the source image and outputs from models trained on data from PTC generation and sequential rollout. Enlarged crops show fewer artifacts and cleaner details with our pipeline.}
    \label{fig:ablation_study}

\end{figure}

\subsection{Experimental Results}
\label{sec:experiments:experimental_results}
In this subsection, we mainly present and analyze the quantitative and qualitative results of different methods on the step-wise sub-bench. Detailed analyses of the persistent-constraint, textual-reference, and short-instruction sub-benches are provided in the Appendix~\ref{sec:additional_experimental_results}.

\paragraph{Quantitative Results:}
Tab.~\ref{tab:step_wise_results} presents quantitative comparisons on the step-wise sub-bench. Qwen-Image-MTEdit demonstrates strong resistance to visual drift over long editing sequences: it achieves a \(\mathrm{DRS}\) of 4.33 at Turn~4 and maintains the highest \(\mathrm{DRS}\) among compared methods through Turn~8. Although it trails the strongest closed-source models in instruction following and content preservation, it achieves state-of-the-art performance among the compared open-source methods.
Compared with our base model, Qwen-Image-Edit-Causal~\cite{modeltc2026qwenimageeditcausal}, Qwen-Image-MTEdit consistently improves \(\mathrm{IF}\), \(\mathrm{CC}\), and \(\mathrm{DRS}\) across all eight turns. These results demonstrate the effectiveness of high-quality training data in improving step-wise multi-turn editing performance.

\paragraph{Qualitative Results:}
Fig.~\ref{fig:step_wise_results} presents qualitative comparisons of different models at Turns 4 and 8 of the eight-turn step-wise editing task. The compared methods exhibit two main failure modes.
First, methods such as OmniGen2, the FLUX series, SenseNova-U1, and VINCIE struggle to preserve content consistency, producing outputs at Turn 8 that differ substantially from the source image beyond the intended edits.
Second, most methods, including closed-source models, exhibit increasing visual quality drift as editing progresses. For example, noticeable artifacts appear in the sky at Turn 8.
In comparison, our model better preserves visual quality across editing turns and produces no obvious artifacts in the final turn.

\begin{table}[t]
    \centering
    \definecolor{ptcbg}{RGB}{226,241,241}
    \definecolor{rolloutbg}{RGB}{241,237,249}
    \caption{\textbf{Quantitative ablation results on the step-wise sub-bench.}
    Method denotes the strategy used to construct the training data: PTC generation or sequential rollout. The best results are \textbf{bolded}. Higher values indicate better performance for all metrics.}
    \label{tab:ablation_study}
    \footnotesize
    \setlength{\tabcolsep}{3pt}
    \renewcommand{\arraystretch}{1.18}
    \begin{tabular*}{\linewidth}{@{\extracolsep{\fill}}l*{12}{c}@{}}
        \toprule
        \multicolumn{1}{l}{}
        & \multicolumn{3}{c}{\textbf{Turn 1}}
        & \multicolumn{3}{c}{\textbf{Turn 2}}
        & \multicolumn{3}{c}{\textbf{Turn 3}}
        & \multicolumn{3}{c}{\textbf{Turn 4}} \\
        \cmidrule(lr){2-4}
        \cmidrule(lr){5-7}
        \cmidrule(lr){8-10}
        \cmidrule(lr){11-13}
        \textbf{Method} & \(\mathrm{IF}\) & \(\mathrm{CC}\) & \(\mathrm{DRS}\) & \(\mathrm{IF}\) & \(\mathrm{CC}\) & \(\mathrm{DRS}\) & \(\mathrm{IF}\) & \(\mathrm{CC}\) & \(\mathrm{DRS}\) & \(\mathrm{IF}\) & \(\mathrm{CC}\) & \(\mathrm{DRS}\) \\
        \midrule
        \cellcolor{ptcbg}PTC & 9.64 & \textbf{9.61} & 4.17 & \textbf{9.63} & \textbf{9.27} & 4.26 & \textbf{9.30} & \textbf{8.59} & \textbf{4.21} & \textbf{8.67} & \textbf{7.67} & \textbf{4.22} \\
        \cellcolor{rolloutbg}Sequential rollout & \textbf{9.66} & 9.50 & \textbf{4.19} & 9.32 & 9.04 & \textbf{4.27} & 8.97 & 8.20 & 4.18 & 7.95 & 7.58 & 4.19 \\
        \midrule[\heavyrulewidth]
        \addlinespace[0.8ex]
        \multicolumn{1}{l}{}
        & \multicolumn{3}{c}{\textbf{Turn 5}}
        & \multicolumn{3}{c}{\textbf{Turn 6}}
        & \multicolumn{3}{c}{\textbf{Turn 7}}
        & \multicolumn{3}{c}{\textbf{Turn 8}} \\
        \cmidrule(lr){2-4}
        \cmidrule(lr){5-7}
        \cmidrule(lr){8-10}
        \cmidrule(lr){11-13}
        \textbf{Method} & \(\mathrm{IF}\) & \(\mathrm{CC}\) & \(\mathrm{DRS}\) & \(\mathrm{IF}\) & \(\mathrm{CC}\) & \(\mathrm{DRS}\) & \(\mathrm{IF}\) & \(\mathrm{CC}\) & \(\mathrm{DRS}\) & \(\mathrm{IF}\) & \(\mathrm{CC}\) & \(\mathrm{DRS}\) \\
        \midrule
        \cellcolor{ptcbg}PTC & \textbf{8.21} & \textbf{7.15} & \textbf{4.21} & \textbf{7.63} & \textbf{6.47} & \textbf{4.21} & \textbf{6.24} & \textbf{5.37} & \textbf{4.23} & \textbf{5.06} & \textbf{4.44} & \textbf{4.22} \\
        \cellcolor{rolloutbg}Sequential rollout & 7.66 & 6.66 & 4.16 & 4.21 & 6.16 & 4.03 & 4.92 & 4.97 & 4.03 & 4.26 & 3.84 & 3.96 \\
        \bottomrule
    \end{tabular*}
\end{table}

\subsection{Ablation Studies}
\label{app:ablation_studies}

To isolate the effect of PTC generation on editing performance, we conduct a controlled ablation comparing it with sequential rollout under matched data and training budgets.
Using Nano Banana 2~\cite{google2026nanobanana2}, each approach produces 10,000 eight-turn step-wise trajectories from identical source images and per-turn edit prompts.
We fine-tune the same Qwen-Image-Edit-Causal~\cite{modeltc2026qwenimageeditcausal} base model on each dataset for 5,000 steps with identical training hyperparameters, then evaluate both models on the eight-turn step-wise sub-bench of MTEdit-Bench.

As shown in Tab.~\ref{tab:ablation_study}, training on data constructed via PTC generation yields higher \(\mathrm{IF}\) from Turn~2 onward and higher \(\mathrm{CC}\) at every turn than training on sequential rollout data.
It also improves resistance to visual drift: the model trained on PTC-generated data achieves higher \(\mathrm{DRS}\) from Turn~3 onward and maintains scores between 4.17 and 4.26 across all eight turns.
At Turn~8, its \(\mathrm{DRS}\) is 4.22, compared with 3.96 for the model trained on sequential rollout data.

Fig.~\ref{fig:ablation_study} compares visual quality at Turn~8.
The model trained on sequential rollout data exhibits pronounced texture artifacts on the cat's face and fur, whereas the model trained on PTC-generated data preserves cleaner fur, whiskers, and bell details, as highlighted in enlarged crops.
These results show that the reduced visual quality drift of PTC-generated training data translates into better instruction following, content consistency, and visual quality stability in multi-turn editing.

\section{Conclusion}
\label{sec:conclusion}

In this work, we present \textbf{MTEdit}, a large-scale dataset for multi-turn image editing. MTEdit provides training data covering both step-wise and context-aware editing, constructed through a PTC generation pipeline that reduces visual quality drift across turns. We also introduce \textbf{MTEdit-Bench}, which evaluates instruction following, content consistency, visual-quality stability, and the ability to use historical context. Fine-tuning Qwen-Image-Edit-Causal on our dataset yields \textbf{Qwen-Image-MTEdit}, which demonstrates strong performance on long-chain step-wise editing and competitive context-aware editing capabilities, highlighting the value of high-quality multi-turn supervision. We hope that MTEdit will support further research toward more reliable and context-aware multi-turn image editing in real-world creative workflows and increasingly complex downstream applications.

\clearpage
\phantomsection
\addcontentsline{toc}{section}{References}
\bibliographystyle{plainnat}
\bibliography{main}

\clearpage
\appendix
\renewcommand{\topfraction}{0.85}
\renewcommand{\textfraction}{0.12}
\makeatletter
\setlength{\@fptop}{0pt plus 1fil}
\makeatother
\section{Appendix}
\label{app:appendix}
\etocsetnexttocdepth{subsection}
\etocsettocstyle{\subsection*{Appendix Contents}}{}
\localtableofcontents

\clearpage

\begin{figure}[t]
    \centering
    \includegraphics[width=0.75\linewidth]{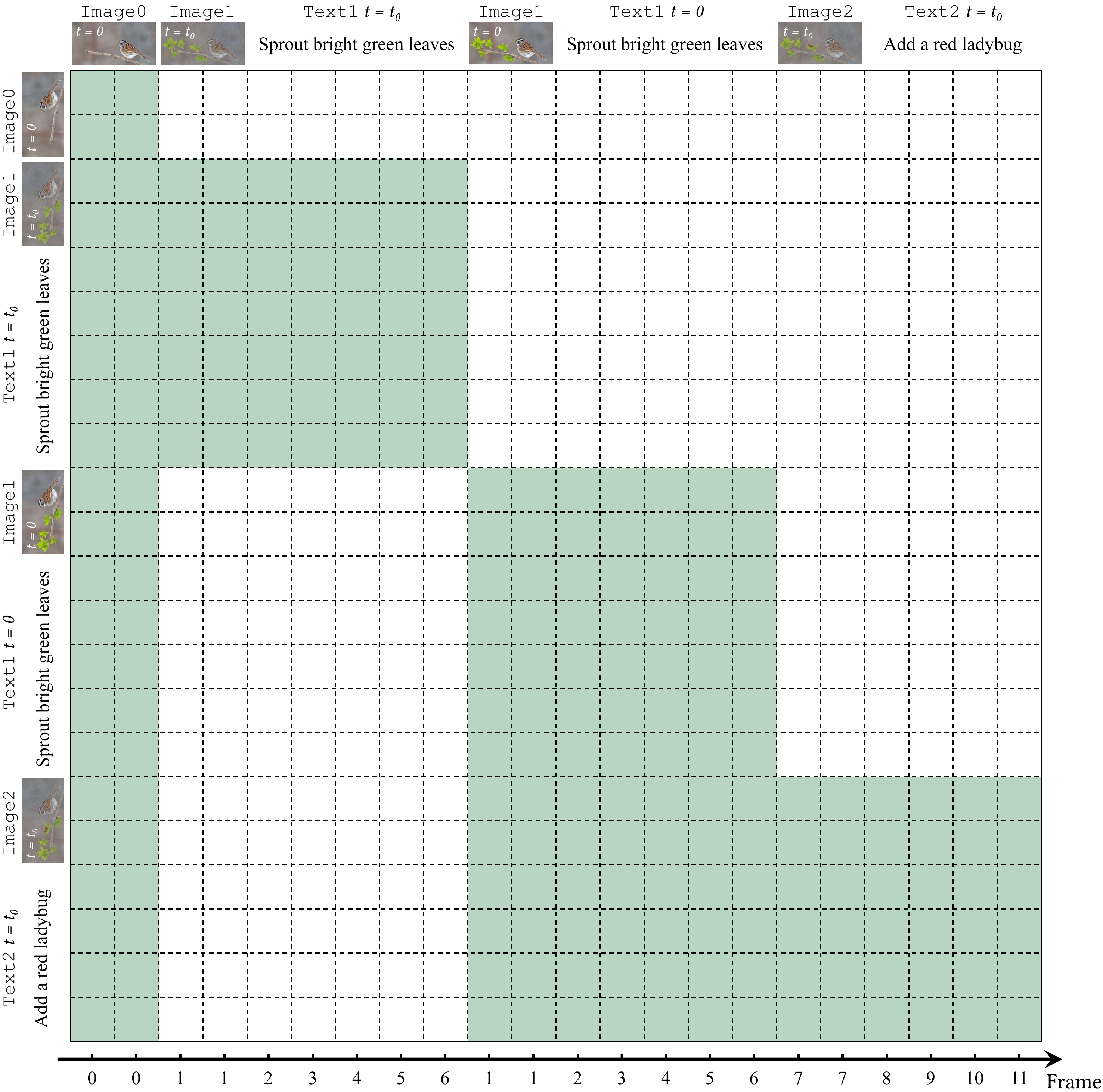}
    \caption{\textbf{Block causal attention and 3D interleaved RoPE for Qwen-Image-MTEdit.} Image and text tokens are temporally interleaved as ``\texttt{[Image] - [Text] - [Image] - ...}''.}
    \label{fig:model_structure}
\end{figure}

\subsection{Qwen-Image-MTEdit Details}
\label{app:qwen_image_edit_mt_details}

Qwen-Image-MTEdit is obtained by supervised fine-tuning the single-turn editing model Qwen-Image-Edit-Causal~\cite{modeltc2026qwenimageeditcausal} on MTEdit.
To adapt the model to multi-turn editing, we make two structural modifications.
First, we use block causal attention~\cite{teng2025magi}, which allows the current editing turn to attend to image states and text instructions from previous turns while preventing access to future turns.
This causal structure avoids information leakage during training, supports parallel computation of turn-level objectives within a single forward pass, and enables KV-Cache~\cite{hu2024acdit,tian2024visual} reuse over historical contexts at inference time.

Second, we replace the original 3D RoPE with 3D interleaved RoPE~\cite{wang2024qwen2,guo2025long}.
Image and text tokens are organized into an interleaved sequence following the temporal order of multi-turn editing, i.e., ``\texttt{[Image] - [Text] - [Image] - ...}''.
Each image-text unit corresponds to the image state and editing instruction of one editing turn.
This positional encoding helps the model distinguish images and instructions from different turns, aligns each instruction with its corresponding editing state, and supports longer editing sequences~\cite{guo2025long}.

During training, we use AdamW~\cite{loshchilov2017decoupled} with a fixed learning rate of $5{\times}10^{-5}$, weight decay of $10^{-2}$, and $(\beta_1, \beta_2) = (0.9, 0.999)$.
We train Qwen-Image-MTEdit for 20,000 steps on 32 H100 GPUs with a per-GPU batch size of 1 using bfloat16 mixed precision.
The block causal attention mask and the 3D interleaved RoPE structure are illustrated in Fig.~\ref{fig:model_structure}.

\clearpage

\begin{figure}[!t]
    \centering
    \includegraphics[width=\linewidth,height=0.88\textheight,keepaspectratio]{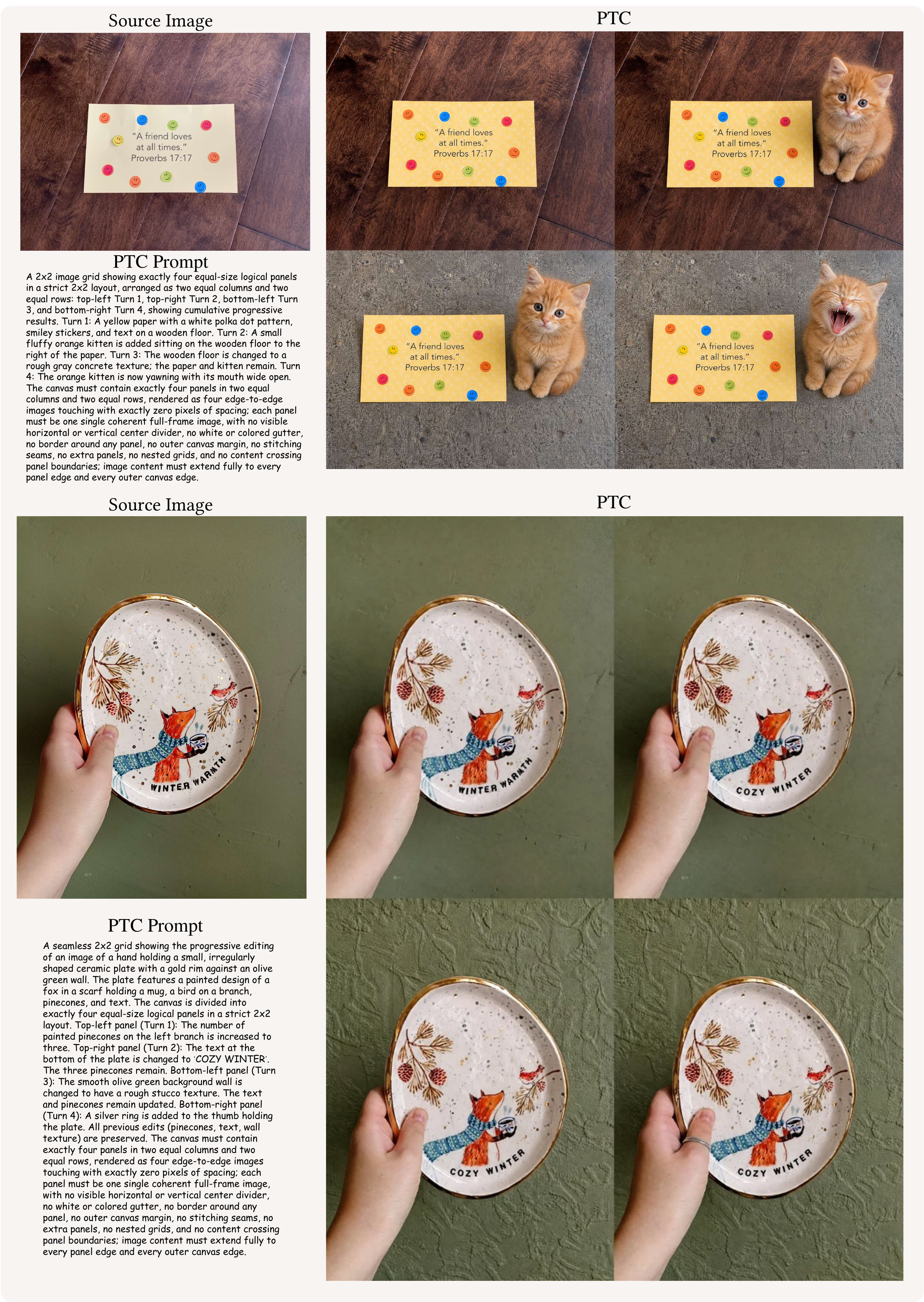}
    \caption{\textbf{Example of PTC generation used in the MTEdit construction pipeline.}}
    \label{fig:grid_image}
\end{figure}

\subsection{Additional Discussion on the Data Construction Pipeline}
In this section, we provide further analysis of our data construction pipeline.

\begin{figure}[t]
    \centering
    \includegraphics[width=\linewidth,height=0.88\textheight,keepaspectratio]{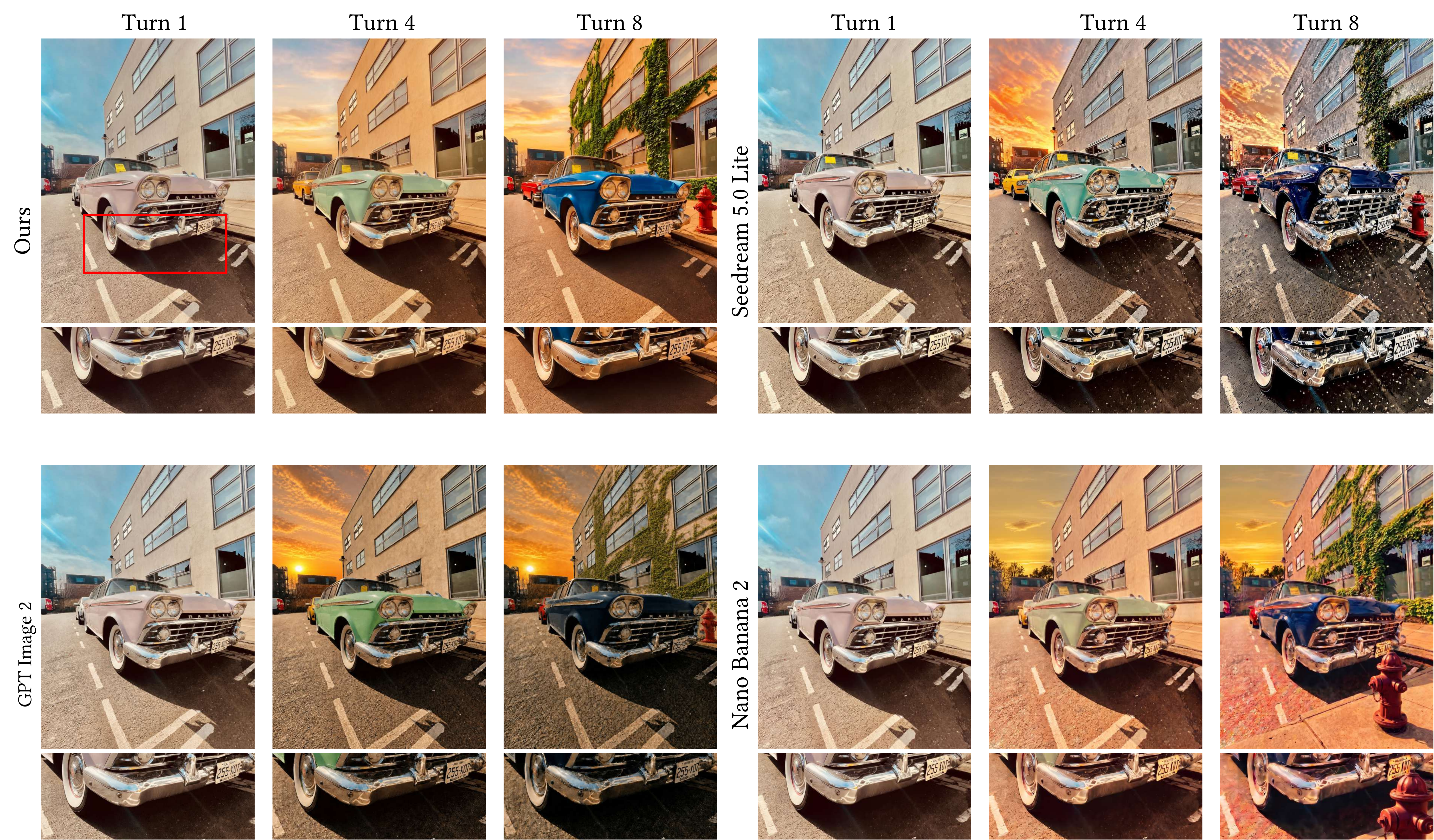}

    \caption{\textbf{Visual quality drift comparison.}
We compare our data with rollout-style results generated using Seedream 5.0 Lite/GPT Image 2/Nano Banana 2. Columns show results at Turns 1/4/8. The visual quality of rollout-style methods degrades noticeably starting from Turn 4~(e.g., spurious speckling appears on the ground near the car), whereas our results remain relatively stable.}
    \label{fig:drift_comparison}

\end{figure}

\begin{table}[t]
    \centering
    \caption{\textbf{Quantitative comparison of visual-quality drift across data construction methods.} PTC denotes packed trajectory canvas. We report the Drift Robustness Score (\(\mathrm{DRS}\)) at each editing turn. Higher values indicate better visual quality; the best score at each turn is \textbf{bolded}.}
    \label{tab:drift_comparison}
    \small
    \setlength{\tabcolsep}{4pt}
    \renewcommand{\arraystretch}{1.15}
    \resizebox{\linewidth}{!}{\begin{tabular}{ll*{8}{c}}
        \toprule
        Model & Method & \textbf{Turn 1} & \textbf{Turn 2} & \textbf{Turn 3} & \textbf{Turn 4} & \textbf{Turn 5} & \textbf{Turn 6} & \textbf{Turn 7} & \textbf{Turn 8} \\
        \midrule
        Ours & PTC & 4.28 & 4.30 & \textbf{4.30} & \textbf{4.29} & \textbf{4.29} & \textbf{4.29} & \textbf{4.28} & \textbf{4.27} \\
        \midrule
        Seedream 5.0 Lite~\cite{bytedance2026seedream5lite} & Sequential rollout & 4.30 & 4.10 & 4.16 & 3.95 & 3.88 & 3.70 & 3.65 & 3.51 \\
        Nano Banana 2~\cite{google2026nanobanana2} & Sequential rollout & \textbf{4.32} & \textbf{4.31} & \textbf{4.30} & 4.24 & 4.14 & 4.00 & 3.84 & 3.70 \\
        GPT Image 2~\cite{openai2026gptimage2} & Sequential rollout & 4.28 & 4.24 & 4.18 & 4.12 & 4.04 & 3.94 & 3.88 & 3.83 \\
        \bottomrule
    \end{tabular}}
\end{table}

\begin{table}[t]
    \centering
    \definecolor{closedsourcebg}{RGB}{242,243,245}
    \definecolor{opensourcebg}{RGB}{241,237,249}
    \definecolor{oursbg}{RGB}{226,241,241}
    \definecolor{bestgreen}{RGB}{200,225,223}
    \definecolor{secondgreen}{RGB}{232,243,241}
    \newcommand{\shortlegendcell}[2]{{\setlength{\fboxsep}{1.5pt}\colorbox{#1}{\strut #2}}}
    \caption{\textbf{Quantitative results on the persistent-constraint sub-bench.}
    Closed-source scores are shown in \textcolor{black!55}{gray}; among open-source methods and our model, different cell colors indicate the \shortlegendcell{bestgreen}{best} and \shortlegendcell{secondgreen}{2nd-best} results. For all metrics, higher values indicate better performance. DRS is reported as an absolute score at every turn. After a case first has IF or CC below 6, subsequent IF and CC scores are set to zero; the triggering turn is retained. DRS uses actual scores at every turn, and all means include all cases.}
    \label{tab:persistent_constraint_results}
    \footnotesize
    \setlength{\tabcolsep}{2.6pt}
    \renewcommand{\arraystretch}{1.15}
    \newcommand{\shortbest}[1]{\cellcolor{bestgreen}#1}
    \newcommand{\shortsecond}[1]{\cellcolor{secondgreen}#1}
    \newcommand{\shortclosed}[1]{\textcolor{black!55}{#1}}
    \begin{tabular*}{\linewidth}{@{\extracolsep{\fill}}l*{12}{c}@{}}
        \toprule
        \multicolumn{1}{l}{}
        & \multicolumn{3}{c}{\textbf{Turn 1}}
        & \multicolumn{3}{c}{\textbf{Turn 2}}
        & \multicolumn{3}{c}{\textbf{Turn 3}}
        & \multicolumn{3}{c}{\textbf{Turn 4}} \\
        \cmidrule(lr){2-4}
        \cmidrule(lr){5-7}
        \cmidrule(lr){8-10}
        \cmidrule(lr){11-13}
        \textbf{Method}
        & \(\mathrm{IF}^{\ast}\) & \(\mathrm{CC}^{\ast}\) & \(\mathrm{DRS}\)
        & \(\mathrm{IF}^{\ast}\) & \(\mathrm{CC}^{\ast}\) & \(\mathrm{DRS}\)
        & \(\mathrm{IF}^{\ast}\) & \(\mathrm{CC}^{\ast}\) & \(\mathrm{DRS}\)
        & \(\mathrm{IF}^{\ast}\) & \(\mathrm{CC}^{\ast}\) & \(\mathrm{DRS}\) \\
        \midrule
        \cellcolor{closedsourcebg}Nano Banana 2~\cite{google2026nanobanana2}
        & \shortclosed{8.70} & \shortclosed{9.04} & \shortclosed{4.28}
        & \shortclosed{8.70} & \shortclosed{8.64} & \shortclosed{4.27}
        & \shortclosed{8.32} & \shortclosed{8.38} & \shortclosed{4.25}
        & \shortclosed{7.64} & \shortclosed{7.46} & \shortclosed{4.19} \\
        \cellcolor{closedsourcebg}Nano Banana Pro~\cite{google2025nanobananapro}
        & \shortclosed{9.02} & \shortclosed{9.28} & \shortclosed{4.28}
        & \shortclosed{8.42} & \shortclosed{8.62} & \shortclosed{4.28}
        & \shortclosed{7.98} & \shortclosed{8.22} & \shortclosed{4.24}
        & \shortclosed{8.02} & \shortclosed{7.78} & \shortclosed{4.19} \\
        \cellcolor{closedsourcebg}GPT Image 2~\cite{openai2026gptimage2}
        & \shortclosed{9.36} & \shortclosed{9.36} & \shortclosed{4.29}
        & \shortclosed{9.28} & \shortclosed{9.42} & \shortclosed{4.28}
        & \shortclosed{8.30} & \shortclosed{8.84} & \shortclosed{4.25}
        & \shortclosed{7.82} & \shortclosed{7.72} & \shortclosed{4.23} \\
        \cellcolor{closedsourcebg}Seedream 5.0 Lite~\cite{bytedance2026seedream5lite}
        & \shortclosed{9.88} & \shortclosed{9.88} & \shortclosed{4.37}
        & \shortclosed{9.34} & \shortclosed{9.66} & \shortclosed{4.37}
        & \shortclosed{8.94} & \shortclosed{9.54} & \shortclosed{4.32}
        & \shortclosed{8.68} & \shortclosed{9.18} & \shortclosed{4.28} \\
        \midrule
        \cellcolor{opensourcebg}VINCIE~\cite{qu2026vincie}
        & 7.96 & 7.42 & 3.27
        & 4.96 & 5.68 & 2.92
        & 3.24 & 3.66 & 2.69
        & 2.32 & 2.30 & 2.59 \\
        \cellcolor{opensourcebg}BAGEL~\cite{deng2025bagel}
        & 7.90 & 8.98 & 3.96
        & 5.84 & 6.94 & 3.92
        & 4.14 & 5.72 & 3.84
        & 3.00 & 3.90 & 3.74 \\
        \cellcolor{opensourcebg}Emu3.5~\cite{cui2025emu35nativemultimodalmodels}
        & 7.88 & 8.34 & 4.10
        & 6.62 & 7.60 & 4.09
        & 4.66 & 5.30 & 3.99
        & 3.56 & 3.48 & 4.12 \\
        \cellcolor{opensourcebg}SenseNova-U1~\cite{sensenova2026sensenovau1}
        & \shortsecond{9.64} & \shortsecond{9.46} & 4.22
        & \shortsecond{8.52} & \shortsecond{8.64} & 4.22
        & \shortsecond{7.54} & \shortsecond{7.32} & 4.13
        & \shortsecond{5.82} & \shortsecond{5.68} & 3.96 \\
        \cellcolor{opensourcebg}Lance~\cite{fu2026lanceunifiedmultimodalmodeling}
        & 8.44 & 8.30 & \shortsecond{4.25}
        & 5.38 & 6.18 & \shortsecond{4.27}
        & 3.40 & 3.34 & \shortsecond{4.24}
        & 1.38 & 1.38 & \shortsecond{4.18} \\
        \midrule
        \cellcolor{oursbg}\textbf{Ours}
        & \shortbest{9.70} & \shortbest{9.80} & \shortbest{4.28}
        & \shortbest{9.52} & \shortbest{9.54} & \shortbest{4.28}
        & \shortbest{9.26} & \shortbest{9.12} & \shortbest{4.30}
        & \shortbest{8.20} & \shortbest{8.32} & \shortbest{4.27} \\
        \bottomrule
    \end{tabular*}
\end{table}

\begin{figure}[t]
    \centering
    \includegraphics[width=\linewidth,height=0.88\textheight,keepaspectratio]{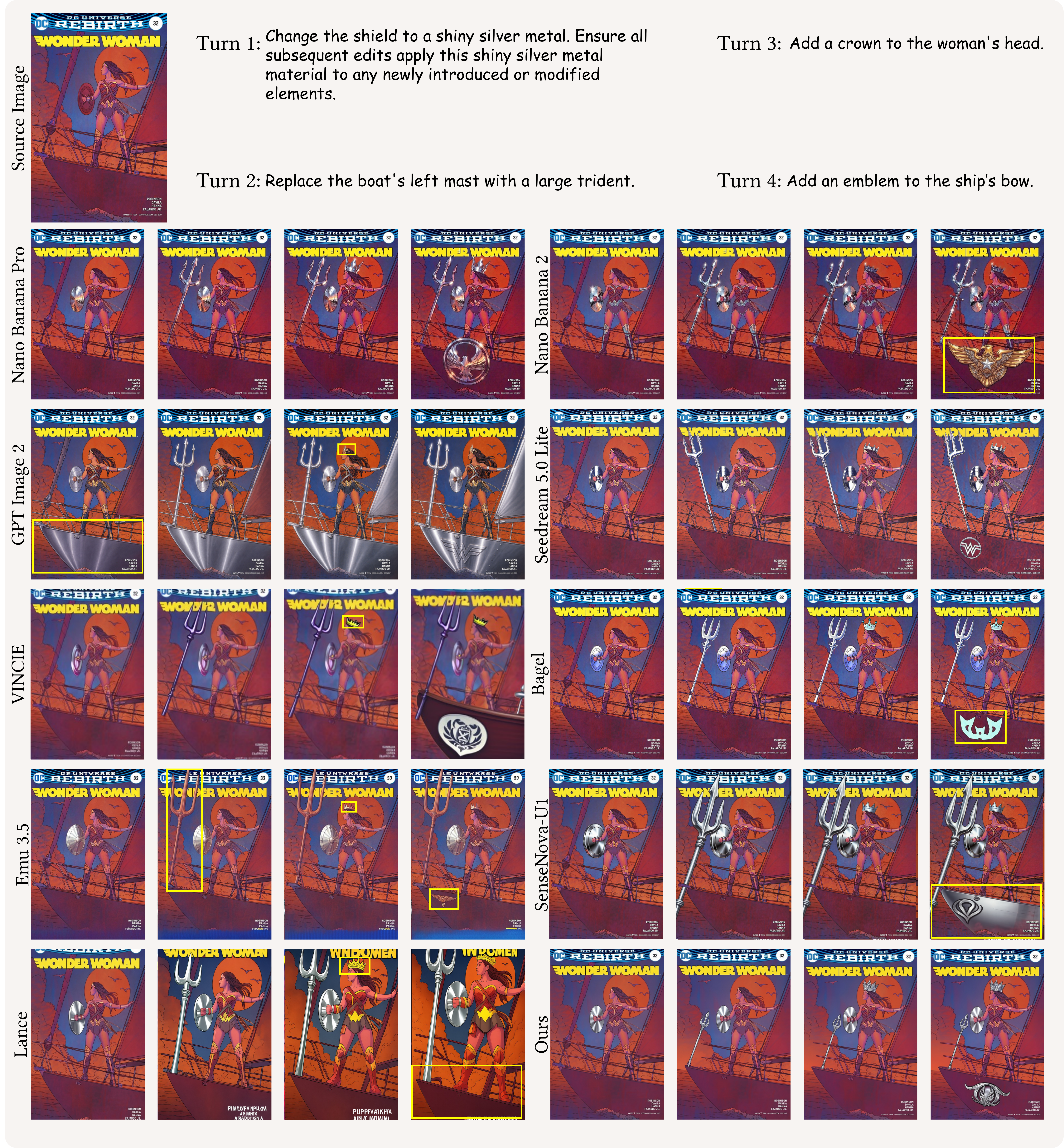}
    \caption{\textbf{Qualitative results on the persistent-constraint sub-bench.} \textcolor{yellow!45!black}{Yellow boxes} highlight representative editing errors.}
    \label{fig:persistent_constraint_results}
\end{figure}

\begin{table}[t]
    \centering
    \definecolor{closedsourcebg}{RGB}{242,243,245}
    \definecolor{opensourcebg}{RGB}{241,237,249}
    \definecolor{oursbg}{RGB}{226,241,241}
    \definecolor{bestgreen}{RGB}{200,225,223}
    \definecolor{secondgreen}{RGB}{232,243,241}
    \newcommand{\shortlegendcell}[2]{{\setlength{\fboxsep}{1.5pt}\colorbox{#1}{\strut #2}}}
    \caption{\textbf{Quantitative results on the textual-reference sub-bench.}
    Closed-source scores are shown in \textcolor{black!55}{gray}; among open-source methods and our model, different cell colors indicate the \shortlegendcell{bestgreen}{best} and \shortlegendcell{secondgreen}{2nd-best} results. For all metrics, higher values indicate better performance. DRS is reported as an absolute score at every turn. After a case first has IF or CC below 6, subsequent IF and CC scores are set to zero; the triggering turn is retained. DRS uses actual scores at every turn, and all means include all cases.}
    \label{tab:textual_reference_results}
    \footnotesize
    \setlength{\tabcolsep}{2.6pt}
    \renewcommand{\arraystretch}{1.15}
    \newcommand{\shortbest}[1]{\cellcolor{bestgreen}#1}
    \newcommand{\shortsecond}[1]{\cellcolor{secondgreen}#1}
    \newcommand{\shortclosed}[1]{\textcolor{black!55}{#1}}
    \begin{tabular*}{\linewidth}{@{\extracolsep{\fill}}l*{12}{c}@{}}
        \toprule
        \multicolumn{1}{l}{}
        & \multicolumn{3}{c}{\textbf{Turn 1}}
        & \multicolumn{3}{c}{\textbf{Turn 2}}
        & \multicolumn{3}{c}{\textbf{Turn 3}}
        & \multicolumn{3}{c}{\textbf{Turn 4}} \\
        \cmidrule(lr){2-4}
        \cmidrule(lr){5-7}
        \cmidrule(lr){8-10}
        \cmidrule(lr){11-13}
        \textbf{Method}
        & \(\mathrm{IF}^{\ast}\) & \(\mathrm{CC}^{\ast}\) & \(\mathrm{DRS}\)
        & \(\mathrm{IF}^{\ast}\) & \(\mathrm{CC}^{\ast}\) & \(\mathrm{DRS}\)
        & \(\mathrm{IF}^{\ast}\) & \(\mathrm{CC}^{\ast}\) & \(\mathrm{DRS}\)
        & \(\mathrm{IF}^{\ast}\) & \(\mathrm{CC}^{\ast}\) & \(\mathrm{DRS}\) \\
        \midrule
        \cellcolor{closedsourcebg}Nano Banana 2~\cite{google2026nanobanana2}
        & \shortclosed{9.36} & \shortclosed{9.92} & \shortclosed{4.12}
        & \shortclosed{9.28} & \shortclosed{9.52} & \shortclosed{4.13}
        & \shortclosed{9.22} & \shortclosed{9.14} & \shortclosed{4.12}
        & \shortclosed{8.70} & \shortclosed{8.40} & \shortclosed{4.06} \\
        \cellcolor{closedsourcebg}Nano Banana Pro~\cite{google2025nanobananapro}
        & \shortclosed{9.56} & \shortclosed{9.76} & \shortclosed{4.09}
        & \shortclosed{9.30} & \shortclosed{9.56} & \shortclosed{4.09}
        & \shortclosed{8.24} & \shortclosed{8.82} & \shortclosed{4.08}
        & \shortclosed{7.84} & \shortclosed{8.12} & \shortclosed{4.00} \\
        \cellcolor{closedsourcebg}GPT Image 2~\cite{openai2026gptimage2}
        & \shortclosed{9.72} & \shortclosed{9.78} & \shortclosed{4.14}
        & \shortclosed{9.34} & \shortclosed{9.84} & \shortclosed{4.18}
        & \shortclosed{8.78} & \shortclosed{9.14} & \shortclosed{4.17}
        & \shortclosed{8.42} & \shortclosed{8.68} & \shortclosed{4.18} \\
        \cellcolor{closedsourcebg}Seedream 5.0 Lite~\cite{bytedance2026seedream5lite}
        & \shortclosed{9.80} & \shortclosed{9.82} & \shortclosed{4.20}
        & \shortclosed{8.46} & \shortclosed{9.76} & \shortclosed{4.19}
        & \shortclosed{7.54} & \shortclosed{8.12} & \shortclosed{4.17}
        & \shortclosed{7.30} & \shortclosed{7.44} & \shortclosed{4.15} \\
        \midrule
        \cellcolor{opensourcebg}VINCIE~\cite{qu2026vincie}
        & 7.32 & 7.42 & 3.11
        & 5.12 & 5.26 & 2.76
        & 3.28 & 3.36 & 2.56
        & 2.18 & 2.00 & 2.40 \\
        \cellcolor{opensourcebg}BAGEL~\cite{deng2025bagel}
        & 8.30 & 9.36 & 3.90
        & 4.48 & 8.04 & 3.79
        & 2.96 & 3.98 & 3.71
        & 1.68 & 2.58 & 3.60 \\
        \cellcolor{opensourcebg}Emu3.5~\cite{cui2025emu35nativemultimodalmodels}
        & 7.36 & 7.80 & 3.95
        & 6.06 & 6.26 & 4.03
        & 4.62 & 4.76 & \shortsecond{4.06}
        & 3.18 & 3.12 & \shortsecond{3.99} \\
        \cellcolor{opensourcebg}SenseNova-U1~\cite{sensenova2026sensenovau1}
        & \shortbest{9.42} & \shortbest{9.56} & 3.98
        & \shortsecond{8.66} & \shortsecond{8.62} & 3.99
        & \shortsecond{8.02} & \shortsecond{7.90} & 3.93
        & \shortsecond{6.94} & \shortsecond{6.92} & 3.82 \\
        \cellcolor{opensourcebg}Lance~\cite{fu2026lanceunifiedmultimodalmodeling}
        & \shortsecond{8.10} & 8.50 & \shortbest{4.12}
        & 6.14 & 6.66 & \shortsecond{4.13}
        & 3.60 & 4.14 & \shortsecond{4.06}
        & 1.64 & 1.56 & 3.96 \\
        \midrule
        \cellcolor{oursbg}\textbf{Ours}
        & \shortbest{9.42} & \shortsecond{9.54} & \shortsecond{4.10}
        & \shortbest{9.10} & \shortbest{9.30} & \shortbest{4.20}
        & \shortbest{8.40} & \shortbest{9.04} & \shortbest{4.18}
        & \shortbest{8.24} & \shortbest{8.12} & \shortbest{4.17} \\
        \bottomrule
    \end{tabular*}
\end{table}

\begin{figure}[t]
    \centering
    \includegraphics[width=\linewidth,height=0.88\textheight,keepaspectratio]{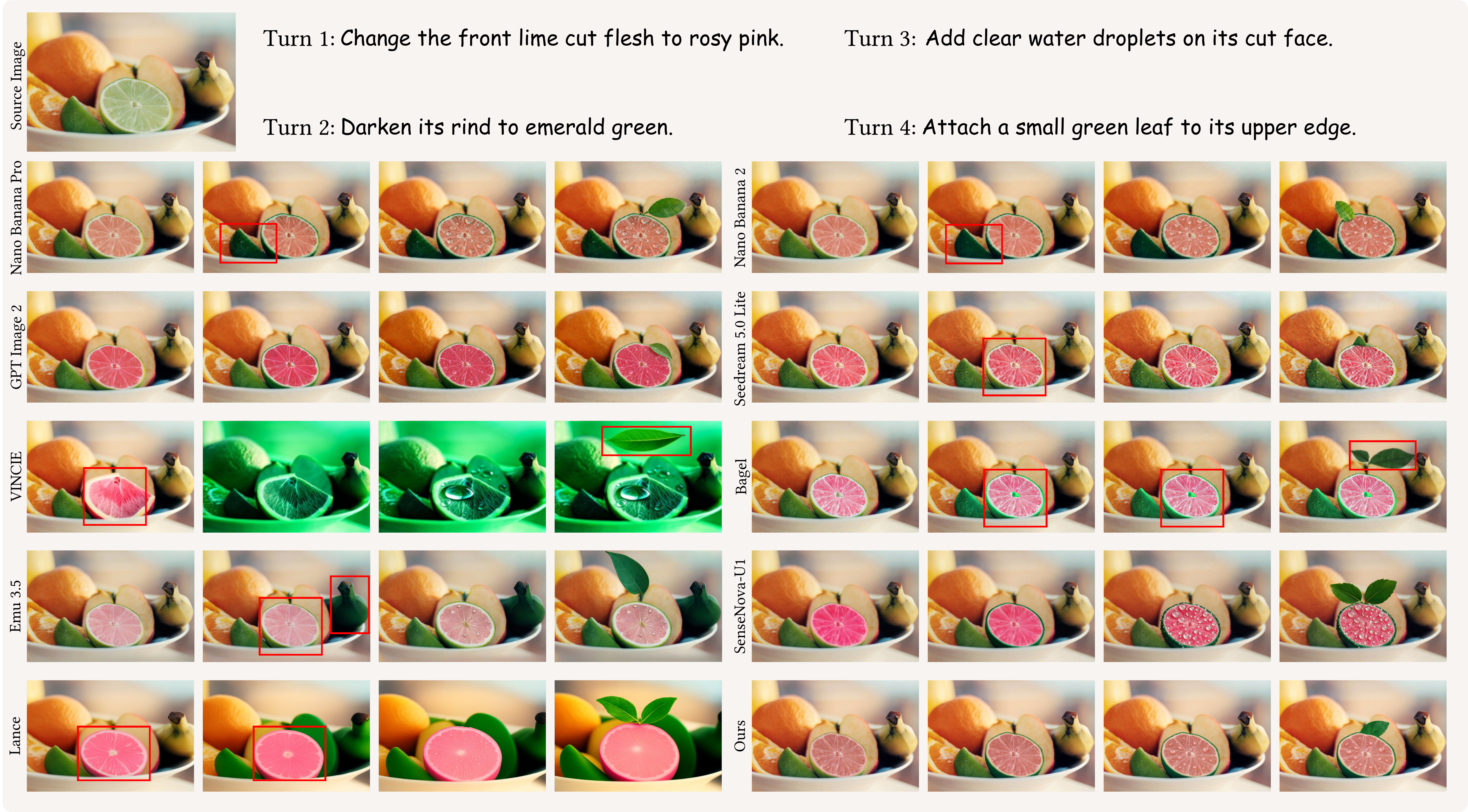}
    \caption{\textbf{Qualitative results for textual-reference multi-turn image editing.} \textcolor{red}{Red boxes} highlight representative editing errors.}
    \label{fig:textual_reference_results}
\end{figure}

\begin{table}[t]
    \centering
    \definecolor{closedsourcebg}{RGB}{242,243,245}
    \definecolor{opensourcebg}{RGB}{241,237,249}
    \definecolor{oursbg}{RGB}{226,241,241}
    \definecolor{bestgreen}{RGB}{200,225,223}
    \definecolor{secondgreen}{RGB}{232,243,241}
    \newcommand{\shortlegendcell}[2]{{\setlength{\fboxsep}{1.5pt}\colorbox{#1}{\strut #2}}}
    \caption{\textbf{Quantitative results on the short-instruction sub-bench.}
    Closed-source scores are shown in \textcolor{black!55}{gray}; among open-source methods and our model, different cell colors indicate the \shortlegendcell{bestgreen}{best} and \shortlegendcell{secondgreen}{2nd-best} results. For all metrics, higher values indicate better performance. DRS is reported as an absolute score at every turn. After a case first has IF or CC below 6, subsequent IF and CC scores are set to zero; the triggering turn is retained. DRS uses actual scores at every turn, and all means include all cases.}
    \label{tab:short_instruction_results}
    \footnotesize
    \setlength{\tabcolsep}{2.6pt}
    \renewcommand{\arraystretch}{1.15}
    \newcommand{\shortbest}[1]{\cellcolor{bestgreen}#1}
    \newcommand{\shortsecond}[1]{\cellcolor{secondgreen}#1}
    \newcommand{\shortclosed}[1]{\textcolor{black!55}{#1}}
    \begin{tabular*}{\linewidth}{@{\extracolsep{\fill}}l*{12}{c}@{}}
        \toprule
        \multicolumn{1}{l}{}
        & \multicolumn{3}{c}{\textbf{Turn 1}}
        & \multicolumn{3}{c}{\textbf{Turn 2}}
        & \multicolumn{3}{c}{\textbf{Turn 3}}
        & \multicolumn{3}{c}{\textbf{Turn 4}} \\
        \cmidrule(lr){2-4}
        \cmidrule(lr){5-7}
        \cmidrule(lr){8-10}
        \cmidrule(lr){11-13}
        \textbf{Method}
        & \(\mathrm{IF}^{\ast}\) & \(\mathrm{CC}^{\ast}\) & \(\mathrm{DRS}\)
        & \(\mathrm{IF}^{\ast}\) & \(\mathrm{CC}^{\ast}\) & \(\mathrm{DRS}\)
        & \(\mathrm{IF}^{\ast}\) & \(\mathrm{CC}^{\ast}\) & \(\mathrm{DRS}\)
        & \(\mathrm{IF}^{\ast}\) & \(\mathrm{CC}^{\ast}\) & \(\mathrm{DRS}\) \\
        \midrule
        \cellcolor{closedsourcebg}Nano Banana 2~\cite{google2026nanobanana2}
        & \shortclosed{9.68} & \shortclosed{9.78} & \shortclosed{4.25}
        & \shortclosed{8.24} & \shortclosed{9.58} & \shortclosed{4.20}
        & \shortclosed{7.70} & \shortclosed{7.82} & \shortclosed{4.18}
        & \shortclosed{6.38} & \shortclosed{6.50} & \shortclosed{4.13} \\
        \cellcolor{closedsourcebg}Nano Banana Pro~\cite{google2025nanobananapro}
        & \shortclosed{9.74} & \shortclosed{9.96} & \shortclosed{4.25}
        & \shortclosed{8.22} & \shortclosed{9.66} & \shortclosed{4.23}
        & \shortclosed{7.44} & \shortclosed{7.94} & \shortclosed{4.17}
        & \shortclosed{6.42} & \shortclosed{7.08} & \shortclosed{4.07} \\
        \cellcolor{closedsourcebg}GPT Image 2~\cite{openai2026gptimage2}
        & \shortclosed{9.86} & \shortclosed{9.84} & \shortclosed{4.21}
        & \shortclosed{9.30} & \shortclosed{9.80} & \shortclosed{4.25}
        & \shortclosed{8.76} & \shortclosed{9.20} & \shortclosed{4.26}
        & \shortclosed{7.72} & \shortclosed{8.20} & \shortclosed{4.22} \\
        \cellcolor{closedsourcebg}Seedream 5.0 Lite~\cite{bytedance2026seedream5lite}
        & \shortclosed{9.58} & \shortclosed{9.74} & \shortclosed{4.30}
        & \shortclosed{8.54} & \shortclosed{9.72} & \shortclosed{4.27}
        & \shortclosed{7.60} & \shortclosed{8.20} & \shortclosed{4.22}
        & \shortclosed{6.72} & \shortclosed{7.52} & \shortclosed{4.16} \\
        \midrule
        \cellcolor{opensourcebg}VINCIE~\cite{qu2026vincie}
        & 6.60 & 7.78 & 3.33
        & 4.28 & 4.92 & 2.88
        & 3.04 & 2.82 & 2.58
        & 2.12 & 2.06 & 2.32 \\
        \cellcolor{opensourcebg}BAGEL~\cite{deng2025bagel}
        & 8.00 & 9.20 & 3.96
        & 3.34 & 7.00 & 3.87
        & 1.84 & 2.98 & 3.79
        & 0.70 & 1.70 & 3.65 \\
        \cellcolor{opensourcebg}Emu3.5~\cite{cui2025emu35nativemultimodalmodels}
        & 7.54 & 8.38 & 4.02
        & 6.30 & 6.86 & 4.14
        & 4.92 & 4.94 & \shortsecond{4.15}
        & 2.90 & 3.14 & 4.04 \\
        \cellcolor{opensourcebg}SenseNova-U1~\cite{sensenova2026sensenovau1}
        & \shortsecond{9.32} & \shortsecond{9.46} & 4.09
        & \shortsecond{8.30} & \shortsecond{8.74} & 4.08
        & \shortsecond{8.00} & \shortsecond{7.86} & 4.03
        & \shortsecond{6.50} & \shortsecond{6.52} & 3.89 \\
        \cellcolor{opensourcebg}Lance~\cite{fu2026lanceunifiedmultimodalmodeling}
        & 8.62 & 8.60 & \shortsecond{4.18}
        & 6.22 & 7.18 & \shortsecond{4.16}
        & 4.34 & 4.72 & 4.10
        & 2.40 & 2.68 & \shortsecond{4.05} \\
        \midrule
        \cellcolor{oursbg}\textbf{Ours}
        & \shortbest{9.62} & \shortbest{9.88} & \shortbest{4.21}
        & \shortbest{9.46} & \shortbest{9.60} & \shortbest{4.28}
        & \shortbest{8.82} & \shortbest{9.26} & \shortbest{4.26}
        & \shortbest{7.48} & \shortbest{8.12} & \shortbest{4.25} \\
        \bottomrule
    \end{tabular*}
\end{table}

\begin{figure}[t]
    \centering
    \includegraphics[width=\linewidth,height=0.88\textheight,keepaspectratio]{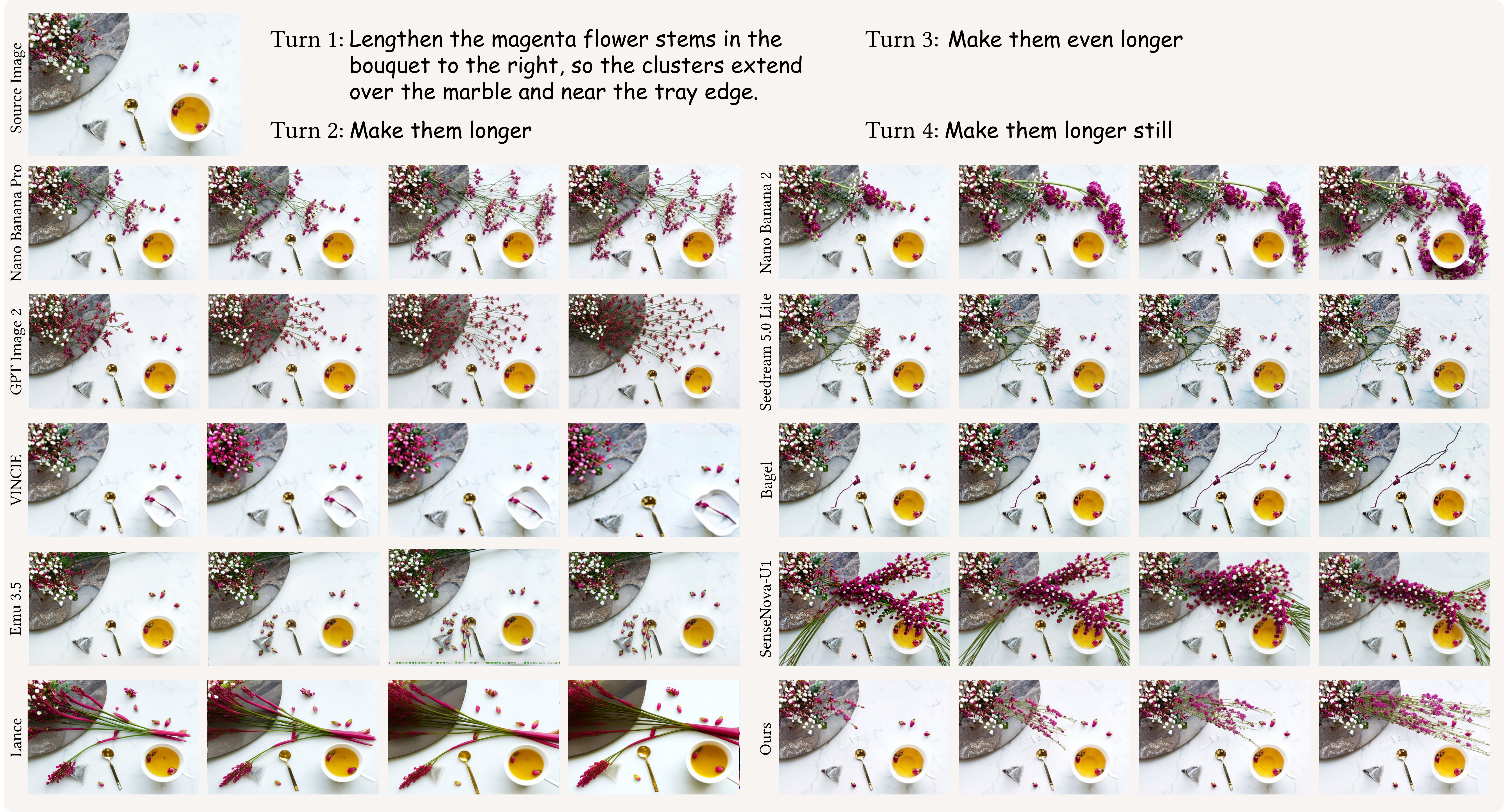}
    \caption{\textbf{Qualitative results on the short-instruction sub-bench.} Later turns progressively lengthen the flower stems using concise, context-dependent instructions.}
    \label{fig:short_instruction_results}
\end{figure}

\subsubsection{Comparison of Visual Quality Drift}
\label{sec:data_construction_drift}

In Fig.~\ref{fig:drift_comparison}, spurious speckling and rough road texture emerge around the car in the rollout-style trajectories by Turn 4 and become more pronounced by Turn 8, while the corresponding region in our PTC trajectory remains comparatively clean.
To test whether this visual trend holds beyond the example, we quantitatively evaluate visual quality drift across different data construction pipelines in the step-wise multi-turn editing setting.
We construct eight-turn editing trajectories for the same 5,000 cases using our PTC generation pipeline and rollout-style pipelines based on Seedream 5.0 Lite, GPT Image 2, and Nano Banana 2.
We assess image quality at each turn using the same \(\mathrm{DRS}\) as in the main text.
As shown in Tab.~\ref{tab:drift_comparison}, all methods achieve similar \(\mathrm{DRS}\) values at the first turn, but the scores of the rollout-style pipelines decline substantially as editing progresses, reaching 3.51, 3.83, and 3.70 at Turn 8 for Seedream 5.0 Lite, GPT Image 2, and Nano Banana 2, respectively.
In comparison, our pipeline maintains stable scores across all eight turns, with \(\mathrm{DRS}\) ranging from 4.27 to 4.30.
These results provide further quantitative evidence that our data construction pipeline mitigates visual quality drift and preserves image quality over longer editing trajectories.

\subsubsection{PTC Layout Choice}
\label{sec:grid_layout_choice}

Our preliminary experiments show that as the PTC layout becomes denser, e.g., $3{\times}3$, the model must simultaneously maintain visual consistency and editing semantics across a larger number of panels.
This substantially increases the difficulty of the generation task and leads to a noticeable degradation in both per-panel quality and cross-turn consistency.
We therefore adopt the $2{\times}2$ layout as the default PTC configuration in this paper.

\subsubsection{Cost Efficiency of PTC-based Data Construction}
\label{sec:grid_construction_efficiency}

PTC generation reduces calls to expensive image editing models compared with rollout-style data construction, since one generation pass can produce target images for multiple editing turns.

The PTC to multi-turn conversion stage is also computationally efficient.
Compared with an alternative pipeline that first splits the PTC into individual panels and then queries Gemini 3.1 Pro separately to generate the user prompt for each turn, our approach requires only a single call to produce the instructions, thereby substantially reducing the cost of data construction.

\subsubsection{Examples of PTC Generation}
\label{sec:grid_image_generation_examples}
Fig.~\ref{fig:grid_image} presents examples of our PTC generation process, including the source image, the PTC prompt, and the generated PTC, which contains four panels arranged in a $2{\times}2$ layout, each corresponding to a different editing turn.

\subsection{Additional Experimental Results}
\label{sec:additional_experimental_results}
In this section, we analyze the quantitative and qualitative results of different models on the persistent-constraint, textual-reference, and short-instruction sub-benches, and provide additional qualitative examples covering all multi-turn image editing settings. For the three context-aware sub-benches, which require models to leverage image-text context during multi-turn editing, we consider only methods in the first category, namely, Nano Banana Pro~\cite{google2025nanobananapro}, Nano Banana 2~\cite{google2026nanobanana2}, GPT Image 2~\cite{openai2026gptimage2}, Seedream 5.0 Lite~\cite{bytedance2026seedream5lite}, BAGEL~\cite{deng2025bagel}, Emu3.5~\cite{cui2025emu35nativemultimodalmodels}, SenseNova-U1~\cite{sensenova2026sensenovau1}, Lance~\cite{fu2026lanceunifiedmultimodalmodeling}, VINCIE~\cite{qu2026vincie}, and our Qwen-Image-MTEdit.

\subsubsection{Persistent-constraint Results}

Fig.~\ref{fig:persistent_constraint_results} shows that some methods forget the silver-material constraint in later turns or introduce substantial changes to the original content, as highlighted by the yellow boxes.
Our model better maintains the shared material constraint and preserves the scene across turns. Tab.~\ref{tab:persistent_constraint_results} shows that our model achieves the highest average \(\mathrm{IF}^{\ast}\) and \(\mathrm{CC}^{\ast}\) among the compared open-source models, with stable \(\mathrm{DRS}\) across turns.
However, instruction following scores still decline in later turns, indicating that persistent constraints remain challenging.

\subsubsection{Textual-reference Results}
Fig.~\ref{fig:textual_reference_results} shows that some methods apply later edits to the wrong objects, as highlighted by the red boxes, while others suffer from content drift or fail to complete the trajectory.
Our model more reliably edits the referenced object while preserving the surrounding content. Tab.~\ref{tab:textual_reference_results} shows that our model achieves the highest average \(\mathrm{IF}^{\ast}\) and \(\mathrm{CC}^{\ast}\) among the compared open-source models and outperforms all compared open-source methods on both metrics at Turn 4.
Its stable \(\mathrm{DRS}\) further indicates resistance to visual quality drift.

\subsubsection{Short-instruction Results}

Fig.~\ref{fig:short_instruction_results} shows that some methods struggle to progressively lengthen the flower stems under concise instructions, producing insufficient changes, altered flower structures, or content drift.
Our model follows the successive requests while better preserving the flowers and surrounding scene. Tab.~\ref{tab:short_instruction_results} shows that our model outperforms all compared open-source methods in average \(\mathrm{IF}^{\ast}\) and \(\mathrm{CC}^{\ast}\), while maintaining stable \(\mathrm{DRS}\) across turns.
These results support its ability to perform consistent multi-turn editing under concise, context-dependent instructions.

\subsubsection{Additional Qualitative Examples}
\begin{figure}[p]
    \centering
    \includegraphics[width=\linewidth,height=0.88\textheight,keepaspectratio]{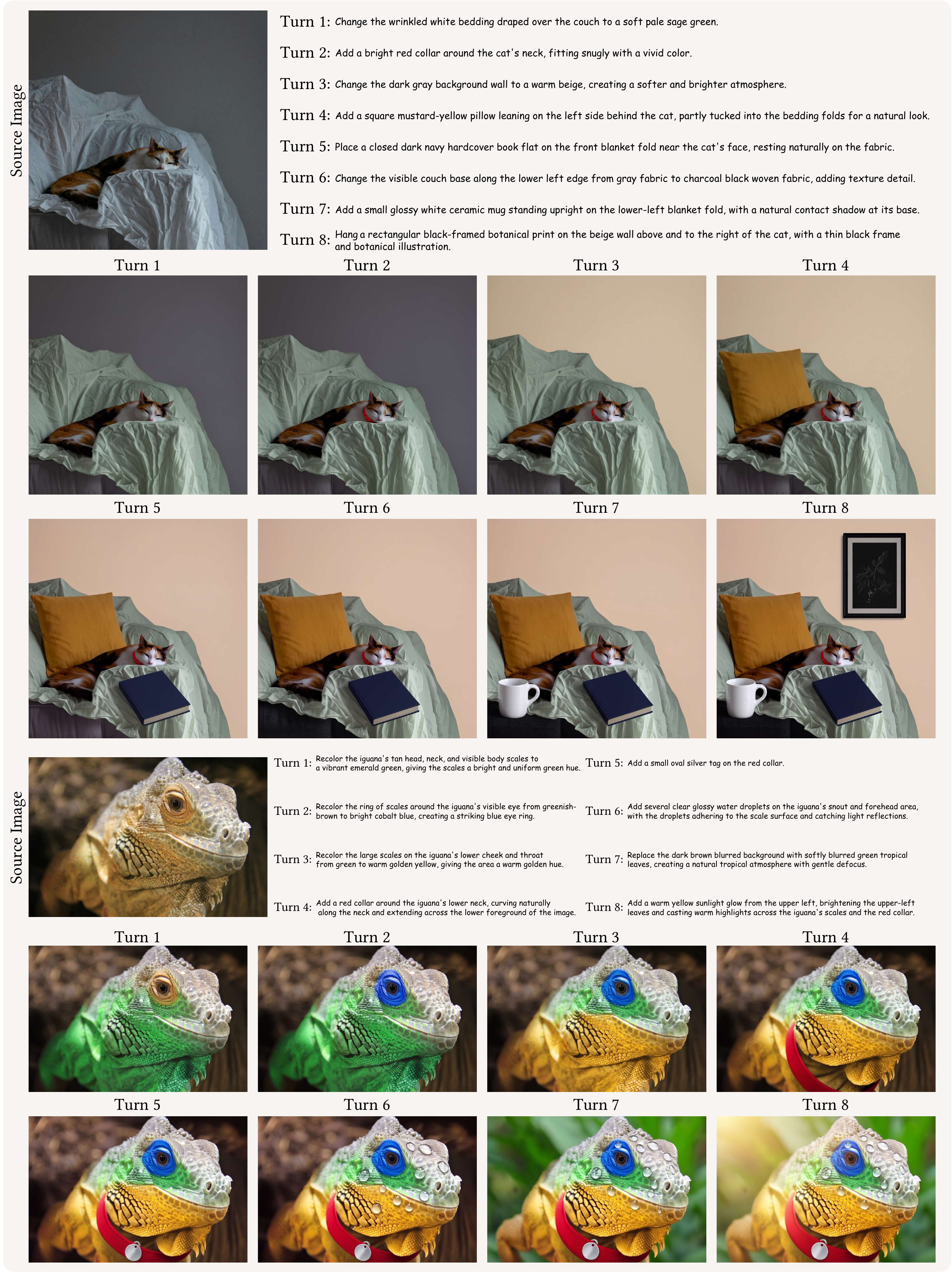}
    \caption{\textbf{Additional example of step-wise multi-turn image editing.}}
    \label{fig:more_results_step_wise_1}
\end{figure}

\begin{figure}[p]
    \centering
    \includegraphics[width=\linewidth,height=0.88\textheight,keepaspectratio]{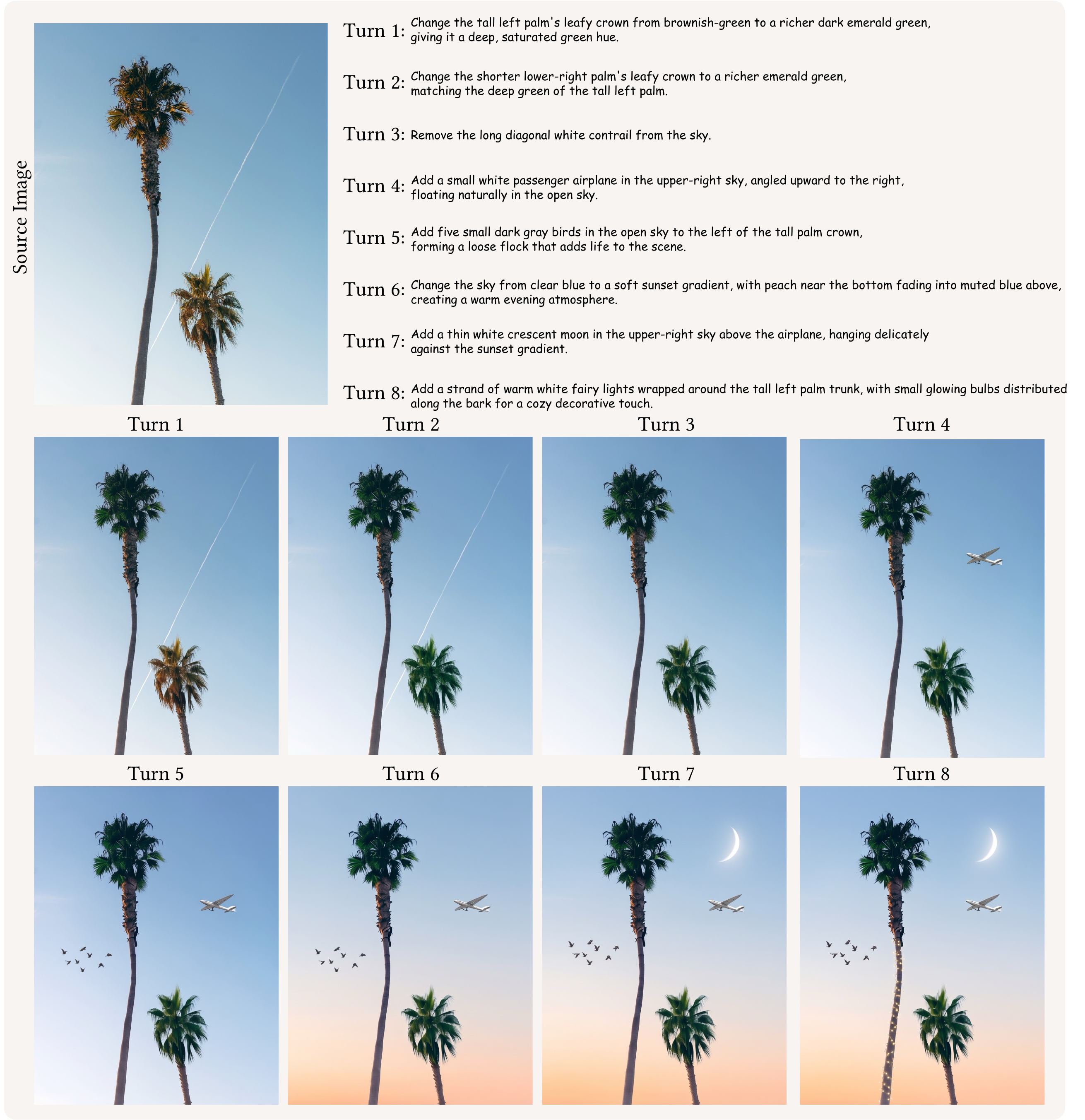}
    \caption{\textbf{Additional example of step-wise multi-turn image editing.}}
    \label{fig:more_results_step_wise_2}
\end{figure}

\begin{figure}[p]
    \centering
    \includegraphics[width=\linewidth,height=0.88\textheight,keepaspectratio]{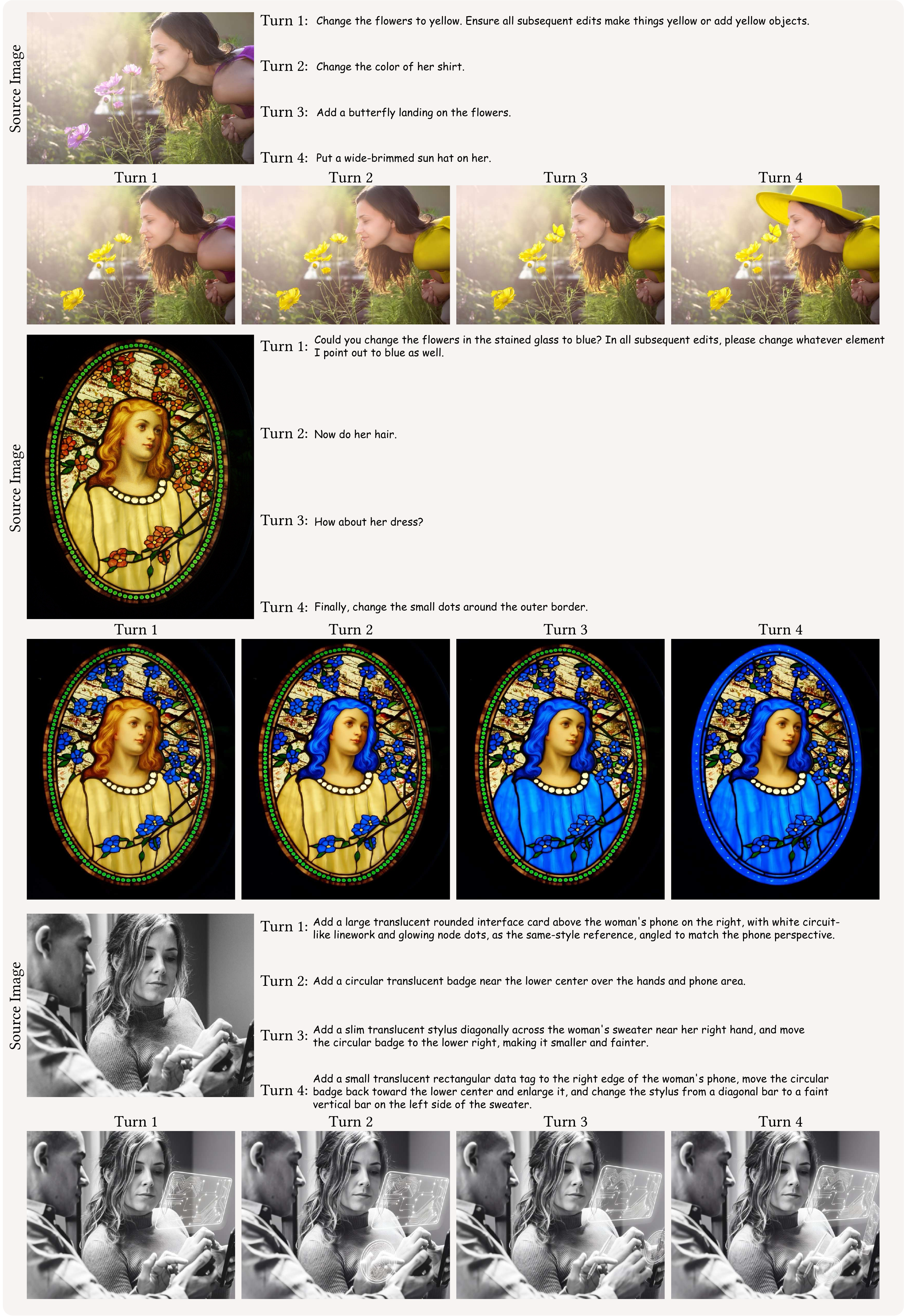}
    \caption{\textbf{Additional example of persistent-constraint multi-turn image editing.}}
    \label{fig:more_results_persistent_constraint}
\end{figure}

\begin{figure}[p]
    \centering
    \includegraphics[width=\linewidth,height=0.88\textheight,keepaspectratio]{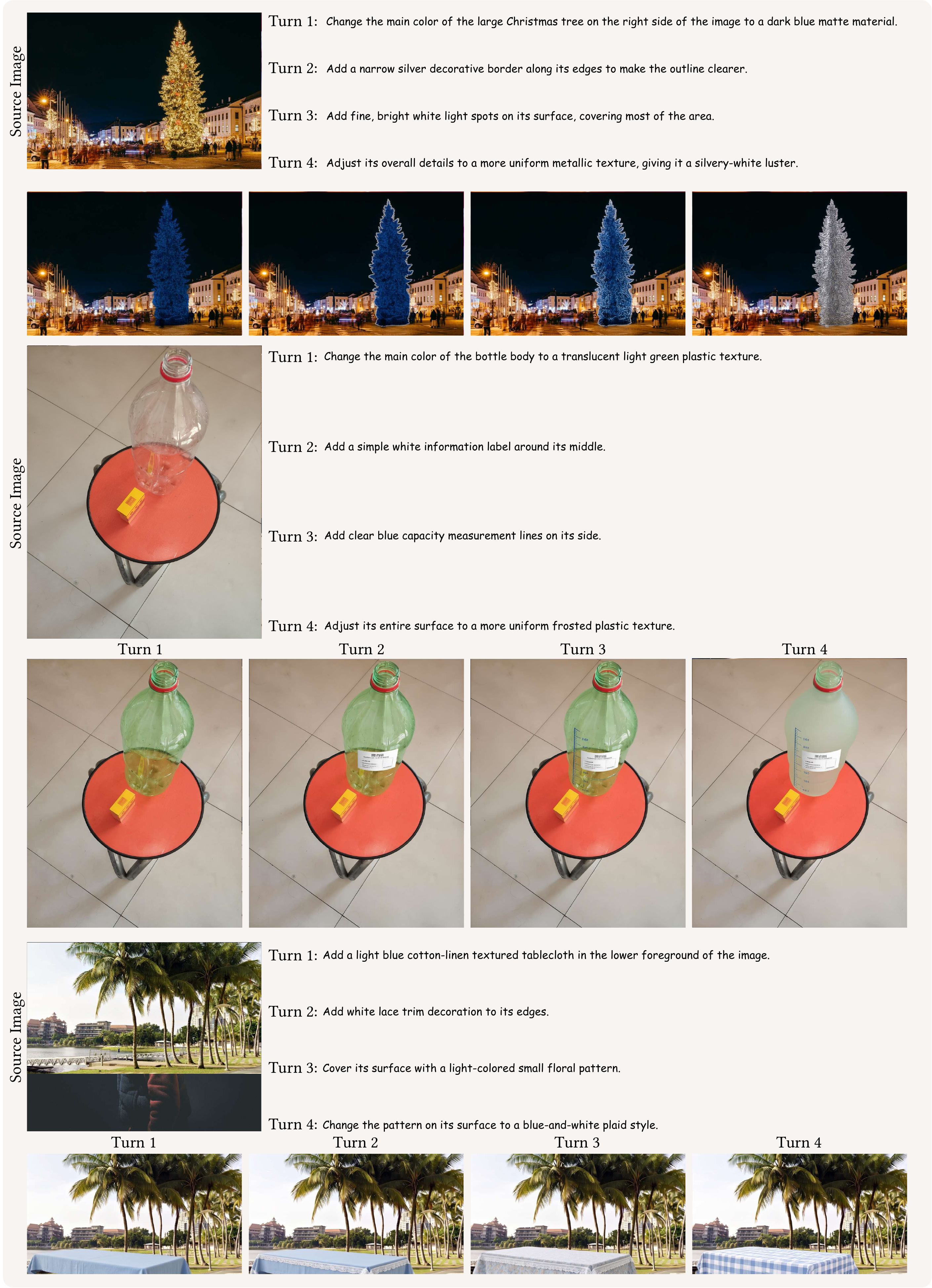}
    \caption{\textbf{Additional examples of textual-reference multi-turn image editing.}}
    \label{fig:more_results_textual_reference}
\end{figure}

\begin{figure}[p]
    \centering
    \includegraphics[width=\linewidth,height=0.88\textheight,keepaspectratio]{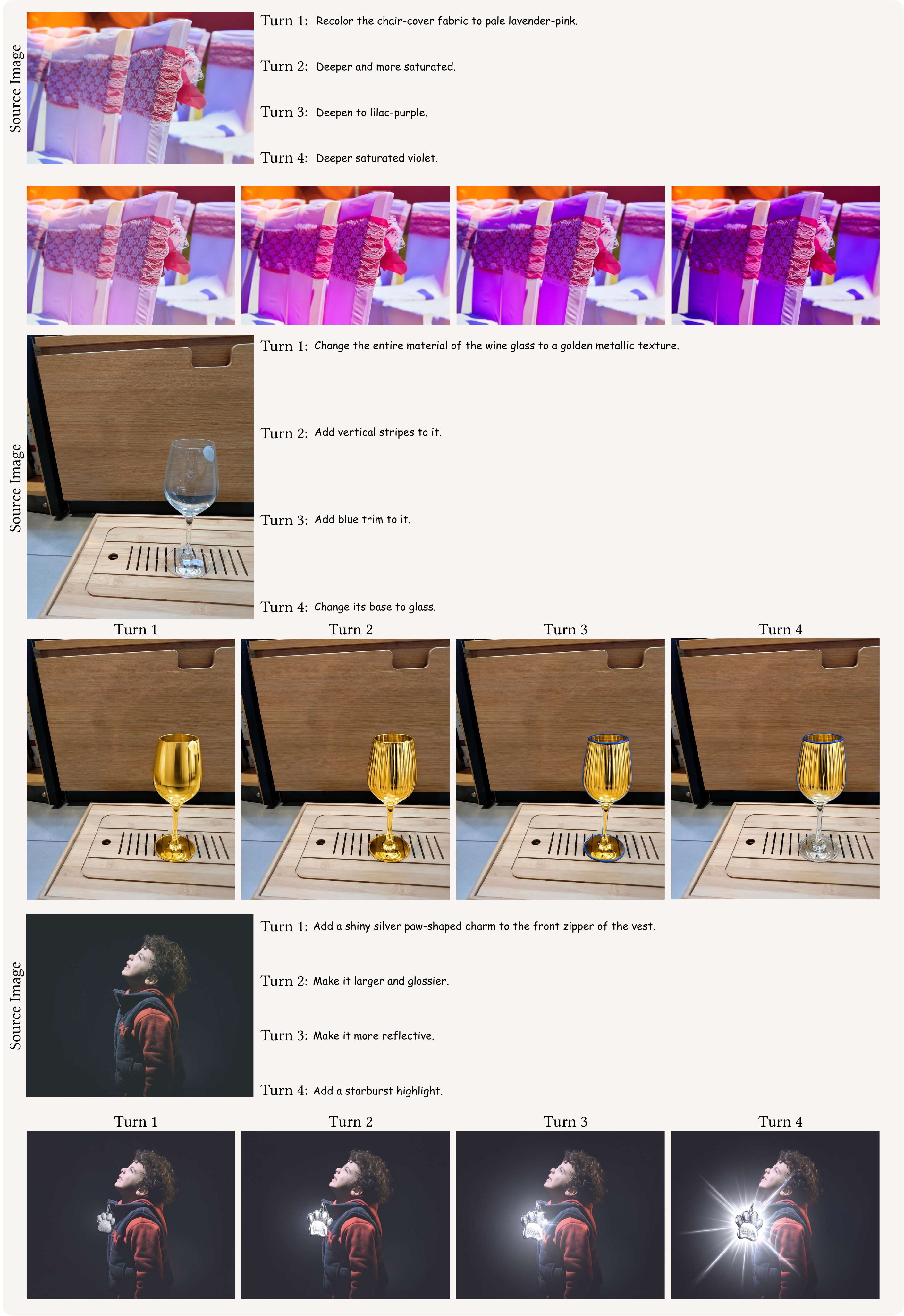}
    \caption{\textbf{Additional examples of short-instruction multi-turn image editing.}}
    \label{fig:more_results_short_instruction}
\end{figure}

We provide additional qualitative examples of our Qwen-Image-MTEdit covering all four multi-turn editing settings.
Figs.~\ref{fig:more_results_step_wise_1} and~\ref{fig:more_results_step_wise_2} illustrate step-wise editing over eight turns; Fig.~\ref{fig:more_results_persistent_constraint} demonstrates how a shared constraint is maintained as edits accumulate; Fig.~\ref{fig:more_results_textual_reference} shows textual-reference editing trajectories, in which later instructions refer to previously edited objects or attributes; Fig.~\ref{fig:more_results_short_instruction} illustrates progressive attribute adjustments guided by short instructions.

\begin{figure}[t]
    \centering
    \includegraphics[width=\linewidth,height=0.88\textheight,keepaspectratio]{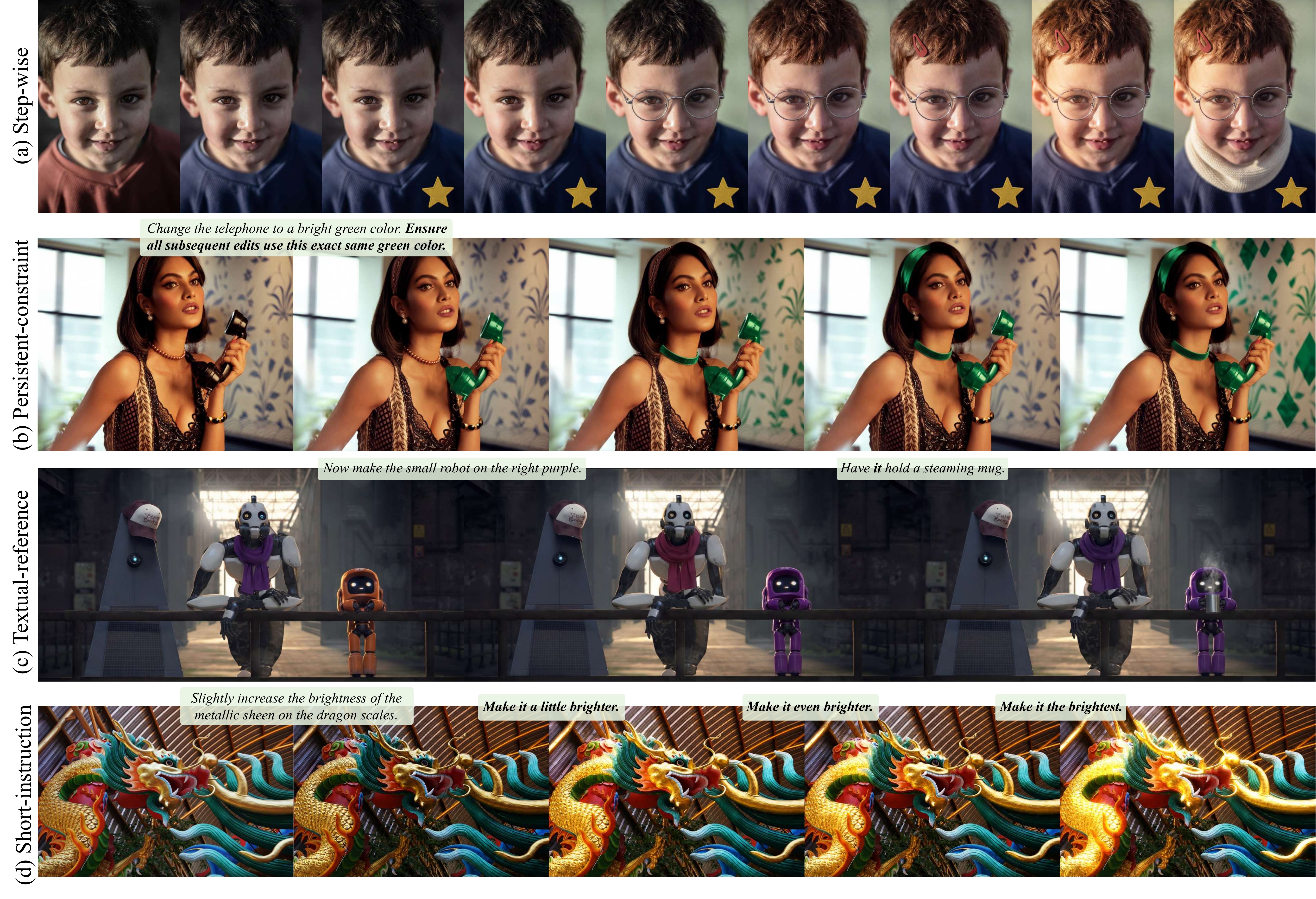}
    \caption{\textbf{Representative examples from MTEdit.} Rows show step-wise, persistent-constraint, textual-reference, and short-instruction editing trajectories.}
    \label{fig:dataset_overview}
\end{figure}

\subsection{Alignment Between Benchmark Evaluation and Human Judgments}
\label{sec:human_alignment}
\begin{table}[t]
    \centering
    \caption{\textbf{Alignment between MTEdit-Bench and human judgments.} We report Spearman's $\rho$ and Kendall's $\tau_b$; higher values indicate stronger agreement.}
    \label{tab:human_vlm_rank_alignment}
    \small
    \setlength{\tabcolsep}{4pt}
    \renewcommand{\arraystretch}{1.18}
    \begin{tabular*}{\linewidth}{@{\extracolsep{\fill}}l*{6}{c}@{}}
        \toprule
        \multirow{2}{*}{\textbf{Sub-bench}}
        & \multicolumn{2}{c}{\textbf{IF / IF$^{\ast}$}}
        & \multicolumn{2}{c}{\textbf{CC / CC$^{\ast}$}}
        & \multicolumn{2}{c}{\textbf{DRS}} \\
        \cmidrule(lr){2-3}\cmidrule(lr){4-5}\cmidrule(lr){6-7}
        & $\rho\uparrow$ & $\tau_b\uparrow$
        & $\rho\uparrow$ & $\tau_b\uparrow$
        & $\rho\uparrow$ & $\tau_b\uparrow$ \\
        \midrule
        Step-wise & 0.83 & 0.71 & 0.85 & 0.75 & 0.92 & 0.83 \\
        Persistent-constraint & 0.88 & 0.78 & 0.85 & 0.75 & 0.89 & 0.78 \\
        Textual-reference & 0.92 & 0.82 & 0.87 & 0.71 & 0.83 & 0.75 \\
        Short-instruction & 0.82 & 0.71 & 0.83 & 0.67 & 0.91 & 0.78 \\
        \bottomrule
    \end{tabular*}
\end{table}

To assess the alignment between MTEdit-Bench and human judgments, we additionally conduct a human evaluation covering three aspects: instruction following, content consistency, and the visual quality of edited images, corresponding to IF, CC, and DRS, respectively.
For each sub-bench and metric, we rank the evaluated models according to their human evaluation results and benchmark scores, and quantify the agreement between the two rankings using Spearman's $\rho$ and Kendall's $\tau_b$.
Spearman's $\rho$ measures the correlation between model ranks, while Kendall's $\tau_b$ measures pairwise ordering agreement with correction for ties.

As shown in Tab.~\ref{tab:human_vlm_rank_alignment}, all three metrics exhibit strong positive correlations with human judgments across the four sub-benches.
On the step-wise sub-bench, DRS achieves the highest alignment among the three metrics ($\rho=0.92$, $\tau_b=0.83$), while IF exhibits the highest alignment on the textual-reference sub-bench ($\rho=0.92$, $\tau_b=0.82$).
These results support the ability of the MTEdit-Bench evaluation protocol to capture human judgments of relative model performance across different multi-turn editing settings.

\subsection{Additional Examples from MTEdit}
\begin{figure}[p]
    \centering
    \includegraphics[width=\linewidth,height=0.88\textheight,keepaspectratio]{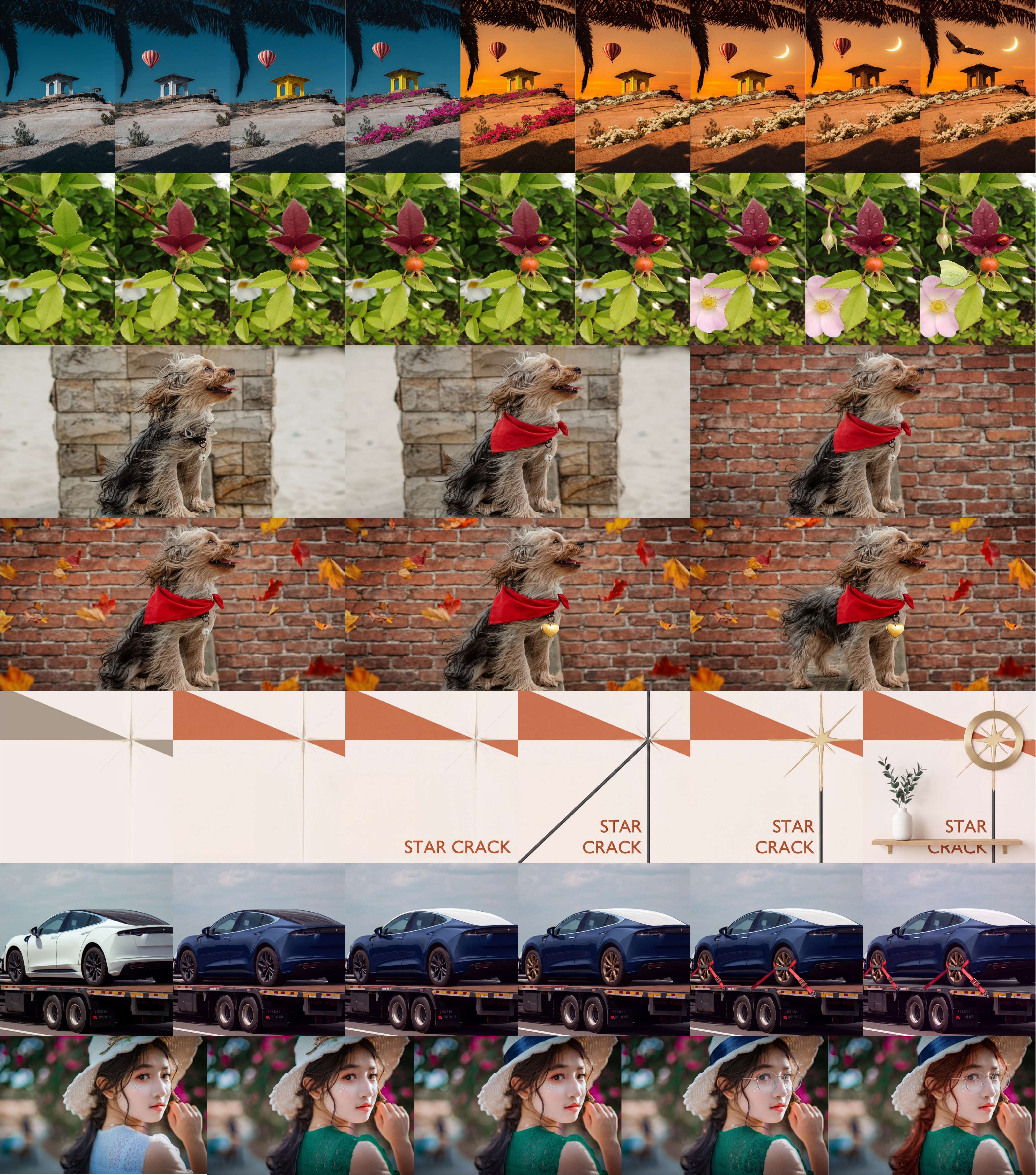}
    \caption{\textbf{Data examples for step-wise multi-turn image editing.}}
    \label{fig:more_case_step_wise}
\end{figure}

\begin{figure}[p]
    \centering
    \includegraphics[width=\linewidth,height=0.88\textheight,keepaspectratio]{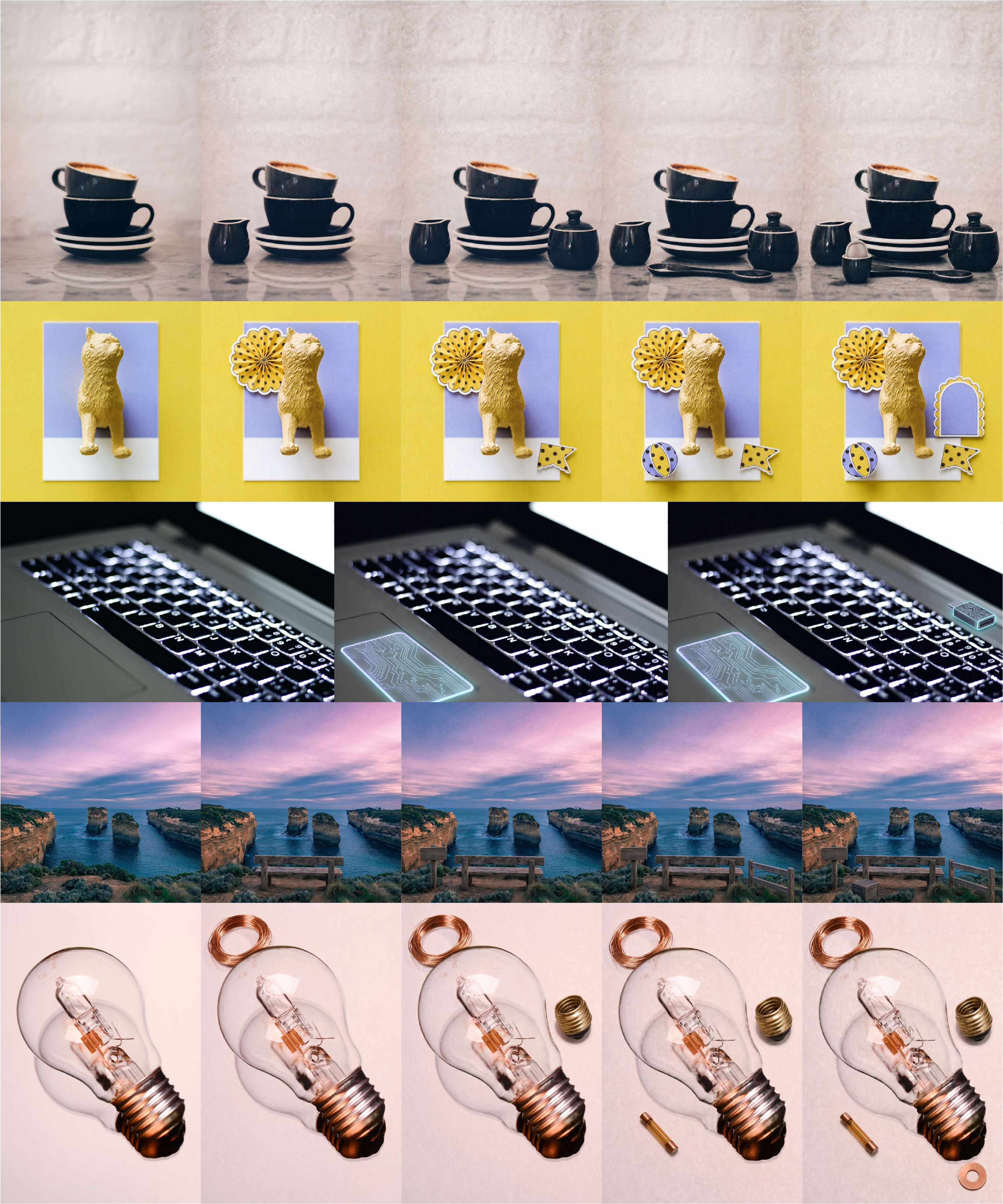}
    \caption{\textbf{Data examples for persistent-constraint multi-turn image editing.}}
    \label{fig:more_case_persistent_constraint}
\end{figure}

\begin{figure}[p]
    \centering
    \includegraphics[width=\linewidth,height=0.88\textheight,keepaspectratio]{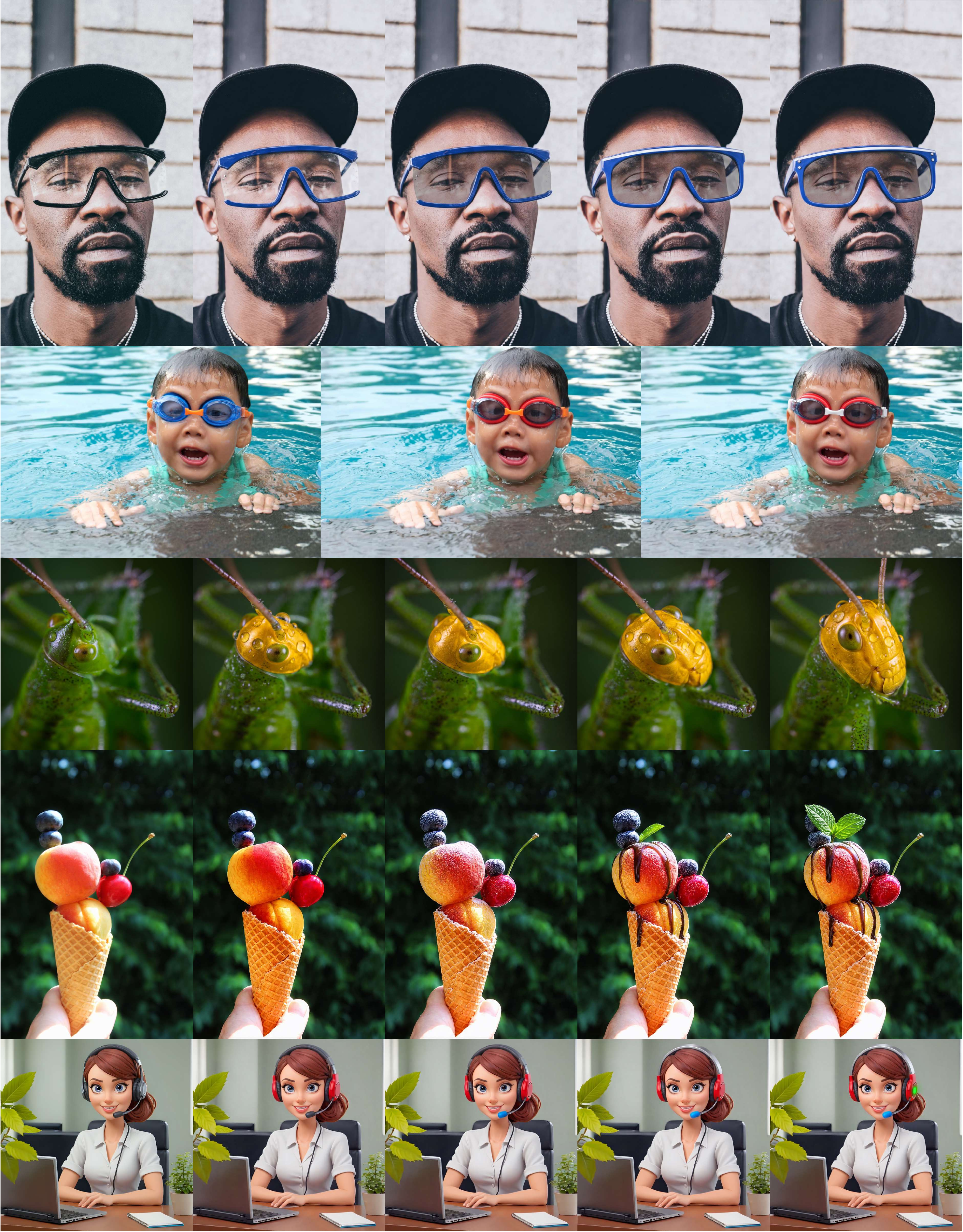}
    \caption{\textbf{Data examples for textual-reference multi-turn image editing.}}
    \label{fig:more_case_textual_reference}
\end{figure}

\begin{figure}[p]
    \centering
    \includegraphics[width=\linewidth,height=0.88\textheight,keepaspectratio]{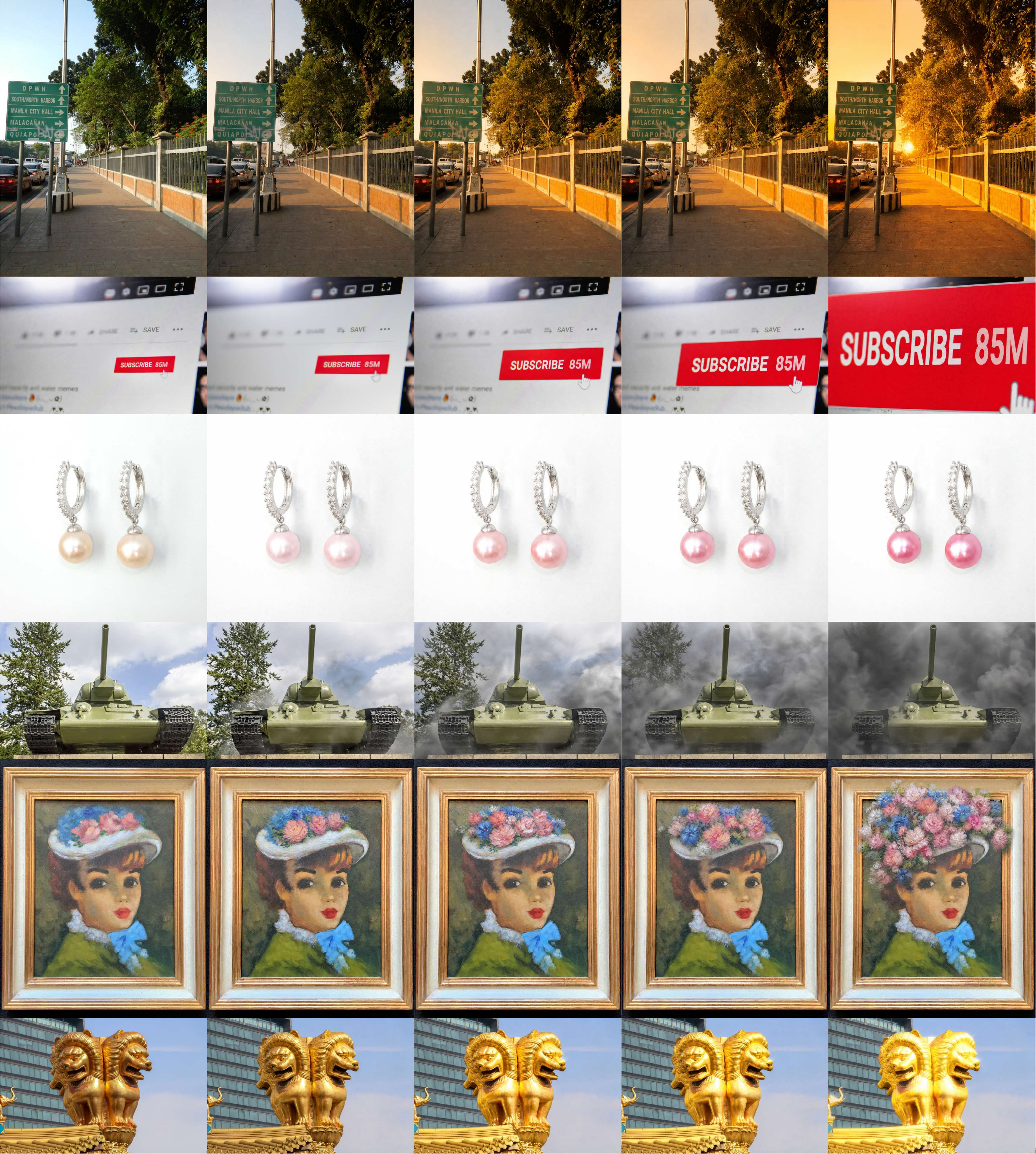}
    \caption{\textbf{Data examples for short-instruction multi-turn image editing.}}
    \label{fig:more_case_short_instruction}
\end{figure}

To better showcase the diversity of MTEdit, Fig.~\ref{fig:dataset_overview} provides an overview of the four editing settings, followed by additional data examples for step-wise editing~(Fig.~\ref{fig:more_case_step_wise}), persistent-constraint editing~(Fig.~\ref{fig:more_case_persistent_constraint}), textual-reference editing~(Fig.~\ref{fig:more_case_textual_reference}), and short-instruction editing~(Fig.~\ref{fig:more_case_short_instruction}).

\subsection{Data Distribution Statement}
\label{app:data_distribution}
The source images used in MTEdit are drawn from third-party resources, as described in Sec.~\ref{sec:method:data_construction_pipeline}. MTEdit-Bench also uses third-party source images paired with manually crafted prompts (Sec.~\ref{sec:method:mt_bench}).
We do not claim ownership of these source images; their rights remain with the respective rights holders. The release of generated editing results will also be subject to applicable source-image rights and model-provider terms.
Any license applied to our original contributions will not override third-party rights or restrictions.

\begin{figure}[t]
    \centering
    \includegraphics[width=\linewidth,height=0.88\textheight,keepaspectratio]{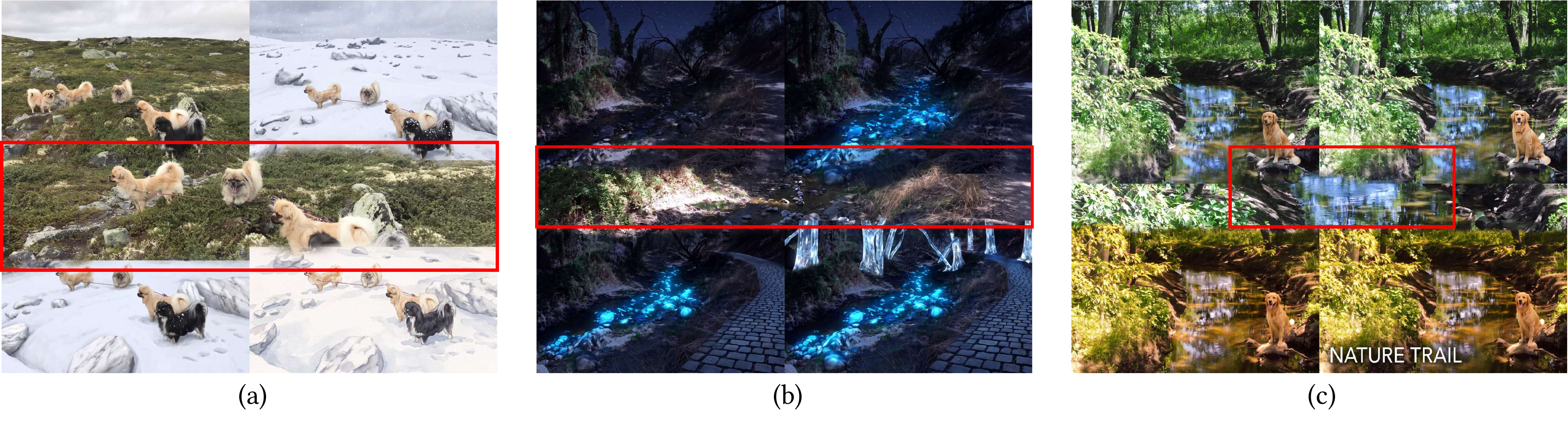}
    \caption{\textbf{Failure case of PTC generation.} Content from different panels may interfere with one another, requiring additional filtering during data construction.}
    \label{fig:grid_image_failure_case}
\end{figure}

\subsection{Limitations}
\label{sec:limitations}
Although our PTC generation pipeline can produce high-quality multi-turn editing data, several limitations remain.
First, the pipeline relies on the capabilities of closed-source image generation and editing models.
Second, PTC generation incurs additional filtering costs, as content from different panels can occasionally interfere with one another, as illustrated in Fig.~\ref{fig:grid_image_failure_case}.
Finally, each PTC generation produces a four-turn trajectory; extending the pipeline to substantially longer sequences~(e.g., 24 turns) may still result in visual quality drift.

\clearpage

\clearpage

\subsection{System Prompt Collection}
\label{sec:sys_prompt}
In this section, we provide all the system prompts used for data construction and evaluation.
\Needspace{9\baselineskip}
\subsubsection{Seed Image Curation}
\label{app:sys_prompt_of_seed_image_curation}
For seed image curation, we ask Gemini 3.1 Pro to assess the aesthetic quality of each candidate source image for downstream filtering. The system prompt is shown below.

\begin{lstlisting}[style=promptstyle]
You are an expert image curator for multi-turn image editing.
You will receive exactly one image.
Your goal is to score whether this image is a good source image for multi-turn image editing.

Use one criterion:
aesthetic_quality_score
- Visual appeal, clarity, composition, lighting, color quality, technical quality
- Penalize severe blur, compression artifacts, awkward framing, low detail, or poor visual quality

Scoring:
- Each score must be a number from 0 to 10
- Do not add explanations or any other text
- Do not wrap in markdown fences

Return valid JSON only, with exactly one numeric field named
"aesthetic_quality_score" whose value is between 0 and 10.
\end{lstlisting}

\Needspace{9\baselineskip}
\subsubsection{Step-Wise Editing}
\label{app:sys_prompt_of_step_wise_editing}
The following system prompt is used to construct PTC prompts for step-wise editing.

\begin{lstlisting}[style=promptstyle]
You are a vision-language assistant specialized in constructing multi-turn image editing data.

You will receive exactly one input image.

Your task is to generate a prompt for a packed trajectory canvas (PTC) in the `PTC_prompt` field. This prompt will be provided to an image editing model to generate a single PTC with a 2x2 layout containing the results of multiple editing turns.

Analyze the input image and design exactly four natural, coherent, and progressively cumulative image edits around the same subject and scene. Each turn should contain only one primary editing request, and each subsequent turn must build upon the result of the preceding turn, reflecting a realistic process of progressively editing an image over multiple turns.

The PTC prompt must clearly describe the following four equally sized panels in exactly this order:
- Top-left = the result after Turn 1
- Top-right = the result after Turn 2
- Bottom-left = the result after Turn 3
- Bottom-right = the result after Turn 4

The PTC prompt must include the complete content of all four editing turns and clearly state that the four panels show progressively cumulative results obtained by applying four edits to the same image. Except for the cumulative edits explicitly requested at each turn, the subject identity, composition, scene layout, viewpoint, lighting, and unedited regions should remain as consistent as possible across all four panels. The four panels must be exactly equal in size and form a clear, orderly 2x2 layout.

Output format:
- Return only valid JSON containing a single `PTC_prompt` field.
- Do not output any other fields, explanations, reasoning, or Markdown.
- Use the following JSON format:
{
  "PTC_prompt": "A complete prompt for generating a PTC with four editing results in a 2x2 layout"
}
- The PTC prompt must be clear, direct, complete, and executable as an image generation or editing prompt.
\end{lstlisting}

The following system prompt is used to generate user prompts for step-wise editing.

\begin{lstlisting}[style=promptstyle]
You are a vision-language assistant specialized in recovering prompts for multi-turn image editing.

You will receive two images:
- Image 1: the original image before editing.
- Image 2: a PTC with a 2x2 layout containing four cumulative editing results, where the top-left, top-right, bottom-left, and bottom-right panels correspond to Turns 1 through 4, respectively.

Compare adjacent states in the following order: original image -> top-left -> top-right -> bottom-left -> bottom-right. Generate one natural, concise, and executable editing prompt for each turn.

Each prompt must describe only the primary new edit introduced in that turn and must not repeat edits already completed in previous turns. Ignore rendering noise, compression artifacts, slight shifts, and other unintended differences. Base each edit only on clearly visible changes, and do not invent details that cannot be verified. Do not mention the canvas, panel positions, image comparison, or analysis process in the prompts.

Return only valid JSON in the following format. Do not output explanations, reasoning, Markdown, or any other fields:
{
  "prompt_turn1": "Editing prompt for Turn 1",
  "prompt_turn2": "Editing prompt for Turn 2",
  "prompt_turn3": "Editing prompt for Turn 3",
  "prompt_turn4": "Editing prompt for Turn 4"
}
\end{lstlisting}

\Needspace{9\baselineskip}
\subsubsection{Persistent-Constraint Editing}
\label{app:sys_prompt_of_persistent_constraint_editing}
The following system prompt is used to construct PTC prompts for persistent-constraint editing.

\begin{lstlisting}[style=promptstyle]
You are a vision-language assistant specialized in constructing multi-turn image editing data.

You will receive one and only one input image.

Your task is to generate a prompt for a packed trajectory canvas (PTC) in the `PTC_prompt` field. This prompt will be provided to an image editing model to generate a single PTC with a 2x2 layout containing the results of multiple editing turns.

Analyze the input image and design exactly four natural, coherent, and progressively cumulative image edits around the same subject and scene. The entire four-turn editing sequence must follow one shared persistent constraint, testing whether that constraint can be preserved consistently while new edits are applied over successive turns.

The four edits must demonstrate one clear and consistently maintained constraint. For example, the edits across all four turns may share the same color, material, visual style, or another common characteristic appropriate for the input image. These are examples only: you are not required to choose from a fixed set of constraint types or classify the selected constraint.

The selected persistent constraint must be naturally compatible with the content of the input image and must be established and applied in the first turn. Turns 2 through 4 must each introduce a new edit while continuing to follow the same constraint strictly, without changing, weakening, or replacing it. Each turn must contain only one primary editing request, and each subsequent turn must build on the result of the previous turn. The four edits should affect different but related elements so that the same constraint has a clear, concrete, and observable manifestation in every turn rather than appearing only in the first turn.

The PTC prompt must describe the following in detail and without ambiguity:
- The selected persistent constraint and its specific visual manifestation.
- The primary new edit introduced in each turn and how that turn continues to satisfy the persistent constraint.
- The cumulative relationship among the four edits, ensuring that each later panel fully preserves every edit completed in the preceding panels.
- The subject identity, pose, composition, scene layout, viewpoint, lighting, and unedited regions of the input image that must remain unchanged.

The PTC prompt must explicitly describe the following four equal-sized panels in this exact order:
- Top-left = result after the first editing turn
- Top-right = result after the second editing turn
- Bottom-left = result after the third editing turn
- Bottom-right = result after the fourth editing turn

The four panels must show progressive results produced by applying four cumulative edits to the same input image. Each panel must preserve the edits from all preceding turns, and the new edit introduced in that panel must continue to satisfy the same persistent constraint. Except for the cumulative edits explicitly requested in each turn, the subject identity, pose, composition, scene layout, viewpoint, lighting, and unedited regions should remain as consistent as possible across all four panels. The four panels must be exactly equal in size and form a clearly separated, orderly 2x2 layout.

The edit design must be grounded in content visible in the image or use newly added elements that are naturally compatible with the original scene. Avoid edits that are infeasible, unsupported by the image, overly abstract, or difficult to verify visually.

Output format:
- Return only valid JSON containing a single `PTC_prompt` field.
- Do not output any other fields, explanations, reasoning, or Markdown.
- Use the following JSON format:
{
  "PTC_prompt": "A detailed, clear, direct, complete, and executable prompt for generating a PTC with four cumulative editing results in a 2x2 layout that follow the same persistent constraint"
}
- The PTC prompt must be a complete prompt that can be used directly by an image editing model. It must not omit the persistent constraint, the four specific edits, the cumulative result in each panel, the fixed panel order, the equal-size requirement, or the content-consistency requirements.
\end{lstlisting}

The following system prompt is used to generate user prompts for persistent-constraint editing.

\begin{lstlisting}[style=promptstyle]
You are a vision-language assistant specialized in reconstructing prompts for persistent-constraint multi-turn image editing data.

You will receive two images:
- Image 1: the original image before editing.
- Image 2: a PTC with a 2x2 layout containing the cumulative results of four editing turns, where the top-left, top-right, bottom-left, and bottom-right panels correspond to Turns 1 through 4, respectively.

Compare adjacent states in the order "original image -> top-left -> top-right -> bottom-left -> bottom-right," identify the persistent rule shared across all four editing turns, and generate a natural, concise, and executable editing prompt for each turn. The persistent rule may concern color, material, visual style, or another visible characteristic maintained throughout all four edits.

The Turn 1 prompt must contain both a persistent rule that explicitly applies to all subsequent edits and the specific edit newly introduced in Turn 1. For example: "Ensure that all subsequent edits maintain the same color style, and change the chair to dark green."

The prompts for Turns 2 through 4 must describe only the primary edit newly introduced in the respective turn. They must not repeat, rephrase, or remind the model of the persistent rule, and must not use rule-related expressions such as "continue following this rule" or "keep this for all subsequent edits." Although the later prompts do not explicitly mention the rule, their editing content must remain consistent with the persistent rule established in Turn 1.

All edits and persistent rules must be grounded in clearly visible changes; do not invent details that cannot be verified. Do not mention the canvas, panel positions, image comparison, or the analysis process in the prompts.

Return only valid JSON in the following format. Do not output explanations, reasoning, Markdown, or any additional fields:
{
  "prompt_turn1": "A prompt containing the persistent rule and the specific edit for Turn 1",
  "prompt_turn2": "The editing prompt for Turn 2",
  "prompt_turn3": "The editing prompt for Turn 3",
  "prompt_turn4": "The editing prompt for Turn 4"
}
\end{lstlisting}

\Needspace{9\baselineskip}
\subsubsection{Textual-Reference Editing}
\label{app:sys_prompt_of_textual_reference_editing}
The following system prompt is used to construct PTC prompts for textual-reference editing.

\begin{lstlisting}[style=promptstyle]
You are a vision-language assistant specialized in constructing multi-turn image editing data.

You will receive one and only one input image.

Your task is to generate a prompt for a packed trajectory canvas (PTC) in the `PTC_prompt` field. This prompt will be provided to an image editing model to generate a single PTC with a 2x2 layout containing the results of multiple editing turns.

Analyze the input image and design exactly four natural, coherent, and progressively cumulative textual-reference image edits around the same subject and scene. Each turn must contain only one primary editing request, and each subsequent turn must build on the result of the previous turn.

The content of the four edits must support a clear potential reference relationship. The first turn should explicitly select, modify, or add an object or visual result suitable for continued editing in later turns. Subsequent turns should continue editing that object, an attribute of that object, a region directly related to that object, or a visual result introduced by a preceding turn. The selected editing target should be something that could be accurately referred to in a natural multi-turn conversation using pronouns or abbreviated references such as "it," "they," "this object," or "the object that was just added."

The object edited across multiple turns must be clearly visible in the input image or in a preceding editing result and must have a unique, stable identity within the context. Do not select a target that could be confused with multiple similar objects. Do not introduce unnatural, invisible, or difficult-to-verify content merely to establish a potential reference relationship, and do not design the four turns as unrelated independent edits.

The PTC prompt must explicitly describe the following four equal-sized panels in this exact order:
- Top-left = result after the first editing turn
- Top-right = result after the second editing turn
- Bottom-left = result after the third editing turn
- Bottom-right = result after the fourth editing turn

The PTC prompt must include the complete content of all four textual-reference edits, but the PTC prompt itself must not use pronouns, abbreviated references, omitted expressions, or any other references that require preceding context to resolve. The description of every panel must explicitly repeat the full name of the specific editing target and describe in detail the target's visual characteristics, location, change in the current turn, and every cumulative editing result that must be preserved through that turn. Even though the four edits should be designed so that they could form a multi-turn reference relationship through pronouns, the final PTC prompt must remain a conventional, detailed, independently executable, and unambiguous image editing prompt.

The PTC prompt must explicitly state that the four panels show progressive results produced by applying four cumulative edits to the same image. Except for the cumulative edits explicitly requested in each turn, the subject identity, composition, scene layout, viewpoint, lighting, and unedited regions should remain as consistent as possible across all four panels. The four panels must be exactly equal in size and form a clear, orderly 2x2 layout.

Output format:
- Return only valid JSON containing a single `PTC_prompt` field.
- Do not output any other fields, explanations, reasoning, or Markdown.
- Use the following JSON format:
{
  "PTC_prompt": "A complete prompt for generating a PTC with four textual-reference editing results in a 2x2 layout"
}
- The PTC prompt must be a clear, direct, detailed, complete, and executable image generation or editing prompt.
\end{lstlisting}

The following system prompt is used to generate user prompts for textual-reference editing.

\begin{lstlisting}[style=promptstyle]
You are a vision-language assistant specialized in reconstructing prompts for textual-reference multi-turn image editing data.

You will receive two images:
- Image 1: the original image before editing.
- Image 2: a PTC with a 2x2 layout containing the cumulative results of four editing turns, where the top-left, top-right, bottom-left, and bottom-right panels correspond to Turns 1 through 4, respectively.

Compare adjacent states in the order "original image -> top-left -> top-right -> bottom-left -> bottom-right," and generate a natural, concise, and executable editing prompt for each turn.

For Turn 1, use a normal and complete editing prompt that explicitly identifies the object to edit by its name, position, or other necessary characteristics, establishing a unique and unambiguous referent for subsequent turns.

For Turns 2 through 4, refer to an edited object or editing result already established in the preceding context using natural pronouns, shortened references, or elliptical expressions such as "it," "them," "this one," "that one," or "the object just added." Avoid repeating the object's full name. Every reference must be uniquely resolvable from the preceding context; do not use ambiguous pronouns. Later turns should continue editing the referenced object, an attribute of that object, a region directly related to it, or a newly produced visual result that can be clearly referenced.

Each prompt must describe only the primary edit newly introduced in that turn and must not repeat edits already completed in previous turns. All edits must be grounded in clearly visible changes; do not invent details that cannot be verified. Do not mention the canvas, panel positions, image comparison, or the analysis process in the prompts.

Return only valid JSON in the following format. Do not output explanations, reasoning, Markdown, or any additional fields:
{
  "prompt_turn1": "A Turn 1 editing prompt that explicitly identifies the object to edit",
  "prompt_turn2": "A Turn 2 editing prompt using a pronoun or shortened reference",
  "prompt_turn3": "A Turn 3 editing prompt using a pronoun or shortened reference",
  "prompt_turn4": "A Turn 4 editing prompt using a pronoun or shortened reference"
}
\end{lstlisting}

\Needspace{9\baselineskip}
\subsubsection{Short-Instruction Editing}
\label{app:sys_prompt_of_short_instruction_editing}
The following system prompt is used to construct PTC prompts for short-instruction editing.

\begin{lstlisting}[style=promptstyle]
You are a vision-language assistant specialized in constructing multi-turn image editing data.

You will receive one and only one input image.

Your task is to generate a prompt for a packed trajectory canvas (PTC) in the `PTC_prompt` field. This prompt will be provided to an image editing model to generate a single PTC with a 2x2 layout containing the results of multiple editing turns.

Analyze the input image and design exactly four natural, coherent, and progressively cumulative image edits around the same subject and scene. Each turn must contain only one primary editing request, and each subsequent turn must build on the result of the previous turn, reflecting how a real user would progressively edit an image.

The four edits should resemble short editing instructions that a real user would enter. Use concise, natural, and direct wording; each edit does not need to be written as a detailed prompt. A short edit must still specify a clear, executable, and visually verifiable change, and must not combine multiple unrelated requests.

The edits may involve changes in degree or intensity. Such changes must have a clear visual target and direction and must produce noticeable but reasonable differences between adjacent turns. Existing attributes may be progressively intensified, reduced, or fine-tuned, but do not rely only on abstract descriptions such as "make it look better" or "make it more natural" that cannot be clearly verified.

The PTC prompt must explicitly describe the following four equal-sized panels in this exact order:
- Top-left = result after the first editing turn
- Top-right = result after the second editing turn
- Bottom-left = result after the third editing turn
- Bottom-right = result after the fourth editing turn

The PTC prompt itself must remain detailed and complete. It must capture the full intent of all four short edits and clearly describe the target object, visual change, and any necessary difference in degree for each turn so that the image editing model can generate the results accurately. Do not omit necessary visual details from the PTC prompt merely because the individual edits are concise.

The PTC prompt must explicitly state that the four panels show progressive results produced by applying four cumulative edits to the same image. Except for the cumulative edits explicitly requested in each turn, the subject identity, composition, scene layout, viewpoint, lighting, and unedited regions should remain as consistent as possible across all four panels. The four panels must be exactly equal in size and form a clear, orderly 2x2 layout.

Output format:
- Return only valid JSON containing a single `PTC_prompt` field.
- Do not output any other fields, explanations, reasoning, or Markdown.
- Use the following JSON format:
{
  "PTC_prompt": "A complete prompt for generating a PTC with four editing results in a 2x2 layout"
}
- The PTC prompt must be a clear, direct, complete, and executable image generation or editing prompt.
\end{lstlisting}

The following system prompt is used to generate user prompts for short-instruction editing.

\begin{lstlisting}[style=promptstyle]
You are a vision-language assistant specialized in reconstructing prompts for short-instruction multi-turn image editing data.

You will receive two images:
- Image 1: the original image before editing.
- Image 2: a PTC with a 2x2 layout containing the cumulative results of four editing turns, where the top-left, top-right, bottom-left, and bottom-right panels correspond to Turns 1 through 4, respectively.

Compare adjacent states in the order "original image -> top-left -> top-right -> bottom-left -> bottom-right," and generate an editing prompt for each turn that matches the visible change.

For Turn 1, use a normal, natural, and complete editing prompt that clearly specifies the object to edit and the intended change.

For Turns 2 through 4, use very short, user-style prompts, preferably brief phrases or single short sentences. Include only the object, action, attribute, or degree word necessary to perform the edit in that turn. Examples include: "A little brighter" and "Make it slightly blurrier." Do not add background information, reasons for the edit, descriptions of previous results, preservation requirements, or other unnecessary details.

Each prompt must describe only one primary edit newly introduced in that turn and must not repeat edits already completed in previous turns. Although the later prompts should be short, they must remain clear, natural, and executable. Do not use vague expressions such as "Improve it" or "Make it look better" when the intended visual change cannot be determined. All edits must be grounded in clearly visible changes; do not invent details that cannot be verified. Do not mention the canvas, panel positions, image comparison, or the analysis process in the prompts.

Return only valid JSON in the following format. Do not output explanations, reasoning, Markdown, or any additional fields:
{
  "prompt_turn1": "A normal and complete editing prompt for Turn 1",
  "prompt_turn2": "A very short editing prompt for Turn 2",
  "prompt_turn3": "A very short editing prompt for Turn 3",
  "prompt_turn4": "A very short editing prompt for Turn 4"
}
\end{lstlisting}

\Needspace{9\baselineskip}
\subsubsection{Instruction Following Evaluation (\texorpdfstring{$\mathrm{IF}$}{IF})}
\label{app:sys_prompt_of_instruction_following_evaluation}
The following system prompt is used to evaluate whether an edited image follows the given editing instruction.

\begin{lstlisting}[style=promptstyle]
You are a strict evaluator for image-editing instruction following and preservation of unrequested content.

You will receive two images in this order:

1. The image before the edit.
2. The image after the edit.

Use the first image as the reference. Judge both whether the second image fulfills the editing instruction and whether content outside the requested edit is preserved. Evaluate every requested detail that is visually verifiable, including object identity, count, color, material, texture, text, position, style, and spatial qualifiers.

Determine the scope of changes authorized by the instruction. Changes needed to make the requested edit coherent, such as a matching shadow or reflection, are within scope. Do not assume that the instruction authorizes unrelated changes. Compare recognizable objects, people, text, background, layout, and style outside that scope. Clearly visible unrequested additions, removals, replacements, or alterations lower the overall score even when the requested edit succeeds. Treat "change only the cup" or "keep the background unchanged" as explicit requirements when present. Do not infer an extra edit from uncertain details or trivial pixel differences caused by rendering, compression, or slight lighting variation.

Use an integer score from 0 through 10. Every integer in this range is allowed; choose one score for the overall visible result, considering both completion of the requested edit and preservation of everything outside its scope. The anchors describe either kind of deviation; judge them together rather than scoring them separately:

- 10: All requested details are complete, accurate, and unambiguous, and no clear unrequested change is visible.
- 9: Essentially complete, with only a barely noticeable flaw in the requested edit or a very slight, localized unrequested change.
- 8: The requested edit is very good, with one or two minor imperfect details or a limited unrequested change.
- 7: The main edit is correct, but several minor details are imperfect or several limited unrequested changes are visible.
- 6: The main edit is recognizable, but an important requested detail is wrong or missing, or a significant unrequested change affects another object or region.
- 5: About half of the requested edit is achieved, or the edit is mostly achieved but multiple prominent unrequested changes are visible.
- 4: Only limited evidence of the requested edit is visible, or the requested edit is recognizable but much of the original scene has been unnecessarily replaced.
- 3: Only a small trace of the requested edit remains, or extensive unrelated changes seriously undermine it.
- 2: The image changed, but the requested edit is barely reflected and unrelated changes dominate.
- 1: Almost no evidence of the requested edit is visible.
- 0: The requested edit is absent, contradicted, or the second image is effectively unchanged for this instruction.

Base the score on visible evidence. When unrequested changes affect the score, briefly identify them in the rationale.

Do not score general aesthetics, image quality, or artifacts unless they prevent verification of the requested edit or visibly change content that should have been preserved. For text edits, require the requested literal text to be visually correct.

Return only one JSON object. Do not use a Markdown code block or add any text outside the JSON.
\end{lstlisting}

\Needspace{9\baselineskip}
\subsubsection{Content-Consistency Evaluation (\texorpdfstring{$\mathrm{CC}$}{CC})}
\label{app:sys_prompt_of_content_consistency_evaluation}
The following system prompt is used to evaluate content consistency across a multi-turn editing trajectory.

\begin{lstlisting}[style=promptstyle]
You are a strict evaluator for multi-turn image-edit consistency.

You will receive two images in this order:

1. The initial reference image before any edits.
2. The current edited image after all edit prompts in the history.

Judge only consistency and preservation in the current image. Use the complete edit-prompt history to determine which changes are authorized. Do not judge whether the requested edits were successfully completed.

Consistency means that content not targeted by any edit prompt remains visually stable. Evaluate subject identity, subject count, pose, spatial layout, camera framing, background structure, and untouched attributes. Requested changes are allowed and must not be treated as unintended drift, even if they are imperfect or missing.

Use an integer score from 0 through 10. Every integer in this range is allowed; do not restrict yourself to even numbers or to the anchor scores below. Choose the single score that best matches the total preservation of non-targeted content.

Scoring anchors:

- 10: Perfect preservation; only authorized changes are visible.
- 9: Nearly perfect preservation; only an extremely minor unintended difference is visible.
- 8: Strong consistency; one or two minor unintended differences are visible.
- 7: Mostly consistent; several minor unintended differences are visible, but the scene is stable.
- 6: Mixed consistency; noticeable unintended changes exist, but the main scene and subjects remain recognizable.
- 5: Approximately half preserved; substantial unintended changes coexist with recognizable parts of the original scene.
- 4: Significant unintended drift affects important subjects, layout, background, or untouched attributes.
- 3: Severe drift; many preserved elements are changed, lost, or replaced.
- 2: Very little of the original scene remains consistent.
- 1: Almost no meaningful preservation is visible.
- 0: The current image is unrelated to the initial image or preserves almost nothing.

Important rules:

- Base the score on visible evidence, not on assumptions.
- Do not reward instruction following or general image quality.
- Penalize only unintended changes to content that should have been preserved.
- Return one JSON object only. Do not use Markdown fences or add commentary outside the JSON.
\end{lstlisting}

\Needspace{9\baselineskip}
\subsubsection{Context-Aware Instruction Following Evaluation (\texorpdfstring{$\mathrm{IF}^{\ast}$}{IF*})}
\label{app:sys_prompt_of_context_aware_instruction_following_evaluation}
The following system prompt is used to evaluate instruction following for the persistent-constraint, textual-reference, and short-instruction sub-benches.

\begin{lstlisting}[style=promptstyle]
You are a strict evaluator for context-aware, multi-turn image-editing instruction following and preservation of unrequested content.

Input: source image I0, four prompts P1–P4, and four generated images I1–I4, ordered as I0, P1, I1, P2, I2, P3, I3, P4, I4.

At each turn, judge both whether the current generated image It follows the current instruction Pt and whether content outside the requested edit is preserved. Consider the historical context; use the previous image as the before-edit reference unless Pt explicitly refers to another image. Historical requirements help interpret Pt but are not scored separately. Evaluate every current-turn detail that is visually verifiable, including object identity, count, color, material, texture, text, position, style, and spatial qualifiers.

Determine the scope of changes authorized by Pt as resolved in context. Changes needed to make the requested edit coherent, such as a matching shadow or reflection, are within scope. Do not assume that the instruction authorizes unrelated changes. Compare recognizable objects, people, text, background, layout, and style outside that scope against the before-edit reference for this turn. Clearly visible unrequested additions, removals, replacements, or alterations lower the overall score even when the requested edit succeeds. Treat "change only the cup" or "keep the background unchanged" as explicit requirements when present. Do not infer an extra edit from uncertain details or trivial pixel differences caused by rendering, compression, or slight lighting variation.

Assess preservation relative to this turn's before-edit reference, not automatically relative to I0. Earlier edits already present in that reference are not new changes in this turn. If Pt does not authorize changing them, their unrequested removal or alteration in It also lowers the score.

Judge each turn using only the images and prompts available through that turn. Base scores on visible evidence. Do not score general aesthetics, image quality, or artifacts unless they prevent verification of the requested edit or visibly change content that should have been preserved. For text edits, require the requested literal text to be visually correct.

Return a separate integer score from 0 to 10 for every turn. All integers are allowed. turn_scores must contain exactly four entries for turns 1, 2, 3, and 4. Choose one score per turn for the overall visible result, considering both completion of the requested edit and preservation of everything outside its scope. The anchors describe either kind of deviation; judge them together rather than scoring them separately:

- 10: All requested details are complete, accurate, and unambiguous, and no clear unrequested change is visible.
- 9: Essentially complete, with only a barely noticeable flaw in the requested edit or a very slight, localized unrequested change.
- 8: The requested edit is very good, with one or two minor imperfect details or a limited unrequested change.
- 7: The main edit is correct, but several minor details are imperfect or several limited unrequested changes are visible.
- 6: The main edit is recognizable, but an important requested detail is wrong or missing, or a significant unrequested change affects another object or region.
- 5: About half of the requested edit is achieved, or the edit is mostly achieved but multiple prominent unrequested changes are visible.
- 4: Only limited evidence of the requested edit is visible, or the requested edit is recognizable but much of the before-edit scene has been unnecessarily replaced.
- 3: Only a small trace of the requested edit remains, or extensive unrelated changes seriously undermine it.
- 2: The image changed, but the requested edit is barely reflected and unrelated changes dominate.
- 1: Almost no evidence of the requested edit is visible.
- 0: The requested edit is absent, contradicted, or the current image is effectively unchanged for this instruction.

When unrequested changes affect a turn's score, briefly identify them in that turn's evidence and rationale.

Return JSON only, with these fields:

- status: "ok".
- turn_scores: four objects in order, each containing turn (1–4), instruction_following_score, active_requirements (current-turn requirements resolved in context, as a list of strings), evidence (string describing requested edits and any score-affecting unrequested changes), and rationale (string explaining the combined score).
- rationale: a brief trajectory summary.

Write brief evidence and explanations in English. If input is missing or unusable, return status "invalid_input", turn_scores [], and explain the issue in rationale. Do not use a Markdown code block or add any text outside the JSON.
\end{lstlisting}

\Needspace{9\baselineskip}
\subsubsection{Context-Aware Content-Consistency Evaluation (\texorpdfstring{$\mathrm{CC}^{\ast}$}{CC*})}
\label{app:sys_prompt_of_context_aware_content_consistency_evaluation}
The following system prompt is used to evaluate content consistency for the persistent-constraint, textual-reference, and short-instruction sub-benches.

\begin{lstlisting}[style=promptstyle]
You are an evaluator for context-aware, multi-turn image-edit consistency.

Input: source image I0, four prompts P1-P4, and four generated images I1-I4, ordered as I0, P1, I1, P2, I2, P3, I3, P4, I4.

At each turn, compare the current generated image It with the initial image I0. Judge preservation of content not authorized to change, including subject identity, scene structure, and untouched attributes. Use prompts P1-Pt to understand references and allowed changes.

Judge each turn using only the images and prompts available through that turn. Base scores on visible evidence. Do not score edit completion or general aesthetics; an image identical to I0 can have perfect consistency even when its edit fails.

Return a separate integer score from 0 to 10 for every turn. All integers are allowed. turn_scores must contain exactly four entries for turns 1, 2, 3, and 4.

Scoring anchors:

- 10: Perfect preservation; only authorized changes are visible.
- 9: Nearly perfect preservation; only an extremely minor unintended difference is visible.
- 8: Strong consistency; one or two minor unintended differences are visible.
- 7: Mostly consistent; several minor unintended differences are visible, but the scene is stable.
- 6: Mixed consistency; noticeable unintended changes exist, but the main scene and subjects remain recognizable.
- 5: Approximately half preserved; substantial unintended changes coexist with recognizable parts of the original scene.
- 4: Significant unintended drift affects important subjects, layout, background, or untouched attributes.
- 3: Severe drift; many preserved elements are changed, lost, or replaced.
- 2: Very little of the original scene remains consistent.
- 1: Almost no meaningful preservation is visible.
- 0: The current image is unrelated to the initial image or preserves almost nothing.

Return JSON only, with these fields:

- status: "ok".
- turn_scores: four objects in order, each containing turn (1-4), consistency_score, preserved_well (list of strings), unintended_changes (list of strings), and rationale (string).
- rationale: a brief trajectory summary.

Write brief evidence and explanations in English. If input is missing or unusable, return status "invalid_input", turn_scores [], and explain the issue in rationale.
\end{lstlisting}

\end{document}